\documentclass[10pt]{article}

\usepackage{style/arxiv}
\usepackage{fontawesome5}
\usepackage{tcolorbox}

\newcommand{\model}{\textsc{SciScore}}
\newcommand{\dataset}{\textsc{Science-T2I}}

\begin{document}

\shorttitle{Science-T2I: Addressing Scientific Illusions in Image Synthesis}
\shortauthor{Li \etal}
\headerright{\textit{CVPR 2025}}

\title{\textsc{Science-T2I:}\\ Addressing Scientific Illusions in Image Synthesis}
\author{
  Jialuo Li$^{1}$ \quad Wenhao Chai$^{2}$ \quad Xingyu Fu$^{3}$ \quad Haiyang Xu$^{4}$ \quad Saining Xie$^{1}$\\[0.3cm]
  $^{1}$New York University \quad $^{2}$University of Washington\\ $^{3}$University of Pennsylvania \quad $^{4}$University of California, San Diego
}
\date{}
\maketitle

\begin{abstract}
Current image generation models produce visually compelling but scientifically implausible images, exposing a fundamental gap between visual fidelity and physical realism. In this work, we introduce~\dataset, an expert-annotated dataset comprising a training set of over 20k adversarial image pairs and 9k prompts across 16 scientific domains and an isolated test set of 454 challenging prompts. Using this benchmark, we evaluate 18 recent image generation models and find that none scores above 50 out of 100 under implicit scientific prompts, while explicit prompts that directly describe the intended outcome yield scores roughly 35 points higher, confirming that current models can render correct scenes when told what to depict but cannot reason from scientific cues to the correct visual outcome. To address this, we develop~\model, a reward model fine-tuned from CLIP-H that captures fine-grained scientific phenomena without relying on language-guided inference, surpassing \texttt{GPT-4o} and experienced human evaluators by roughly 5 points. We further propose a two-stage alignment framework combining supervised fine-tuning with masked online fine-tuning to inject scientific knowledge into generative models. Applying this framework to FLUX.1[dev] yields a relative improvement exceeding 50\% on~\model, demonstrating that scientific reasoning in image generation can be substantially improved through targeted data and alignment.
\end{abstract}

\vspace{-0.2cm}
\begin{center}
\begin{tabular*}{0.8\textwidth}{@{}c@{\hskip 1.2em}l@{\extracolsep{\fill}}l@{}}
\worldwideweb & \textbf{Website} & \href{https://jialuo-li.github.io/Science-T2I-Web/}{{\color{LabBlue}\LinkFont https://jialuo-li.github.io/Science-T2I-Web/}}\\[0.3em]
\github & \textbf{Code} & \href{https://github.com/Jialuo-Li/Science-T2I}{{\color{LabBlue}\LinkFont https://github.com/Jialuo-Li/Science-T2I}}\\[0.3em]
\huggingface & \textbf{Data} & \href{https://huggingface.co/datasets/Jialuo21/Science-T2I-Trainset}{{\color{LabBlue}\LinkFont https://huggingface.co/datasets/Jialuo21/science-t2i}} \\ [0.3em]
\huggingface & \textbf{Model} & \href{https://huggingface.co/Jialuo21/Science-T2I-Flux-SFT-OFT}{{\color{LabBlue}\LinkFont https://huggingface.co/Jialuo21/Science-T2I-Flux-SFT-OFT}}
\end{tabular*}
\end{center}

\section{Introduction}
\label{sec:intro}

\begin{figure}[t]
    \centering
    \includegraphics[width=0.98\linewidth]{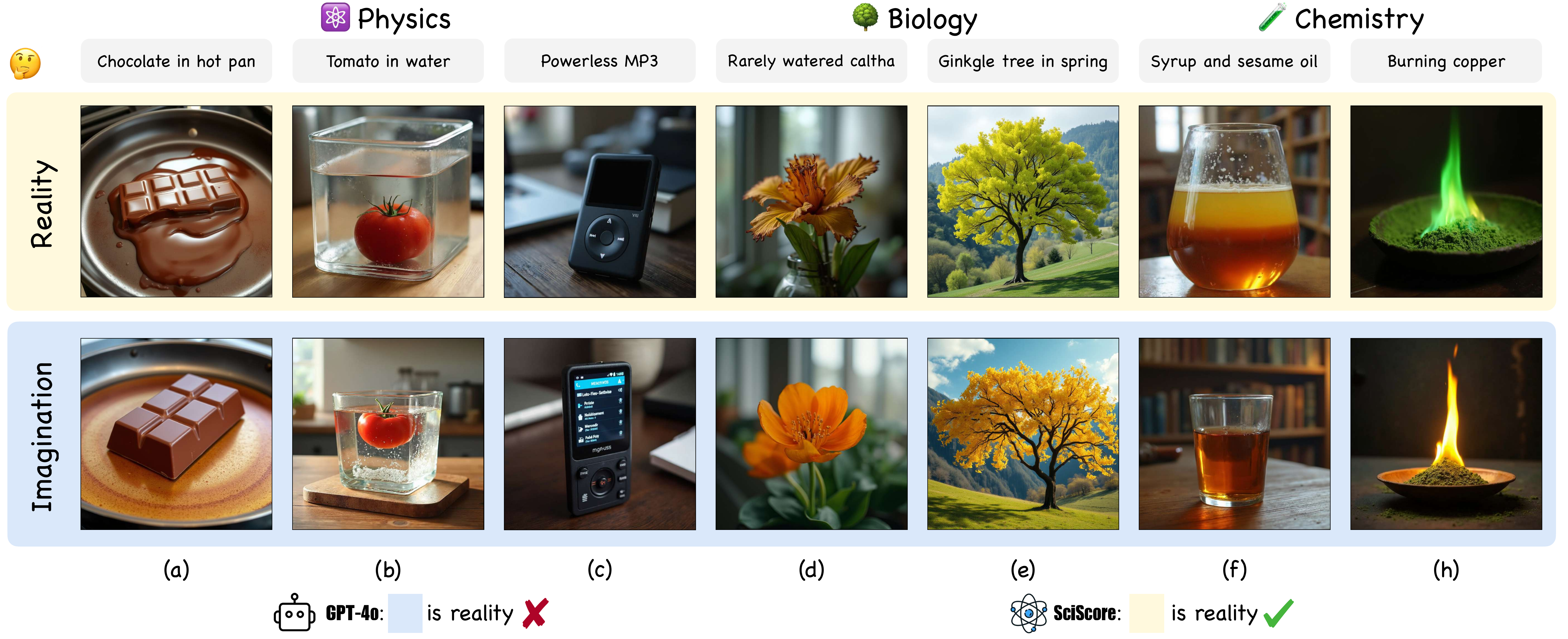}
    \caption{\textbf{Comparison between \texttt{GPT-4o} and~\model.} Given a prompt (in grey) requiring scientific knowledge, FLUX~\cite{FLUX} model generates imaginary images (lower row) that are far from reality (upper row). Moreover, LMMs like \texttt{GPT-4o} \cite{GPT} fail to identify the realistic image, whereas our reward model~\model~succeeds. Notice that the prompts here are summaries of the real prompts that we used for illustration purposes.}
    \label{fig:data_compare}
    \vspace{-6pt}
\end{figure}

The quest to conceptualize the visual world and construct real world simulators has been a longstanding endeavor in the computer vision community~\cite{ferwerda1996model,greenberg1997framework,jensen2001realistic,Zhang_2017_stackgan_ICCV,Chen_2018_SketchyGAN_CVPR,zhang2018stackgan++}. As articulated by~\cite{cohen1993radiosity}, ``The goal of image synthesis is to create, using the computer, a visual experience that is identical to what a viewer would experience when viewing a real environment.'' In alignment with this vision, recent advances in generative modeling have notably improved the performance of image synthesis~\cite{podell2023sdxlimprovinglatentdiffusion,song2022denoisingdiffusionimplicitmodels,rombach2022highresolutionimagesynthesislatent,chen2025multimodal,ye2024learning,cao2023image}.
While these advancements enable the generation of higher resolution, more aesthetically pleasing images with superior Fr\'{e}chet Inception Distance (FID) scores~\cite{FLUX,dalle,podell2023sdxlimprovinglatentdiffusion,yang2024usinghumanfeedbackfinetune}, these models often produce superficial imitations rather than authentic representations of the real visual world~\cite{meng2024phybench,fu2024commonsenset2ichallengetexttoimagegeneration,bansal2024videophy,meng2024phygenbench}.
This limitation often arises from an inadequate understanding of the underlying scientific principles of realism, as demonstrated in the lower row of FLUX~\cite{FLUX} generated images in Figure~\ref{fig:data_compare}. Consequently, the images generated tend to mirror imaginative constructs, resulting in a noticeable gap between these creations and the tangible reality we inhabit.

To bridge this gap between visual imagination and scientific realism, we introduce~\dataset, an expert-annotated dataset partitioned into two isolated components: (1) a training set of over 20k adversarial image pairs and 9k prompts spanning 16 scientific domains across physics, chemistry, and biology, where each pair contrasts a scientifically accurate image with a flawed counterpart to facilitate preference modeling; and (2) a test set of 454 challenging prompts designed to benchmark the scientific plausibility of contemporary image generation models. All data across both splits underwent rigorous validation by human domain experts.

Using this test set, we conduct a systematic evaluation of 18 recent image generation models and find that none achieves a score above 50 out of 100 under implicit prompts. When the same models are evaluated with explicit prompts that directly describe the intended scientific outcome, scores increase by roughly 35 points on average. This gap reveals that current models can render scientifically correct scenes when explicitly instructed, yet cannot reason from implicit scientific cues to the correct visual outcome.

Leveraging the~\dataset~training set, we develop~\model, a reward model that extends CLIP-H~\cite{ilharco_gabriel_2021_5143773} with scientific knowledge to assess whether generated images faithfully reflect the physical principles implied by a prompt. We find that~\model~outperforms large multimodal models (LMMs) such as \texttt{GPT-4o}~\cite{GPT} on this task: while LMMs often overlook fine-grained visual cues, as shown in Figure~\ref{fig:data_compare},~\model~captures them reliably without requiring language-guided inference.

Building upon~\model, we propose a two-stage alignment framework to inject scientific knowledge into generative models. We first apply supervised fine-tuning (SFT) on FLUX.1[dev]~\cite{FLUX} using the~\dataset~training set, then perform online fine-tuning (OFT) with~\model~as the reward signal and a subject-based masking strategy to stabilize optimization. Our contributions are as follows:

\begin{itemize}
    \item We introduce~\dataset, comprising over 20k expert-annotated adversarial pairs for preference alignment and 454 isolated test prompts for benchmarking the scientific realism of image generation models.
    \item Using this benchmark, we evaluate 18 image generation models and find that all score below 50 under implicit prompts, with an average 35-point gap compared to explicit prompts, exposing a systematic lack of scientific reasoning in current systems.
    \item We develop~\model, a reward model that captures subtle scientific phenomena, surpassing state-of-the-art LMMs and human evaluators on our benchmark.
    \item We propose a two-stage alignment framework that improves FLUX.1[dev]~\cite{FLUX} by over 50\% on~\model, demonstrating that scientific reasoning in image generation can be substantially enhanced through targeted data and alignment.
\end{itemize}

\clearpage
\begin{figure}[t]
    \begin{minipage}[b]{0.42\linewidth}
        \includegraphics[width=\linewidth]{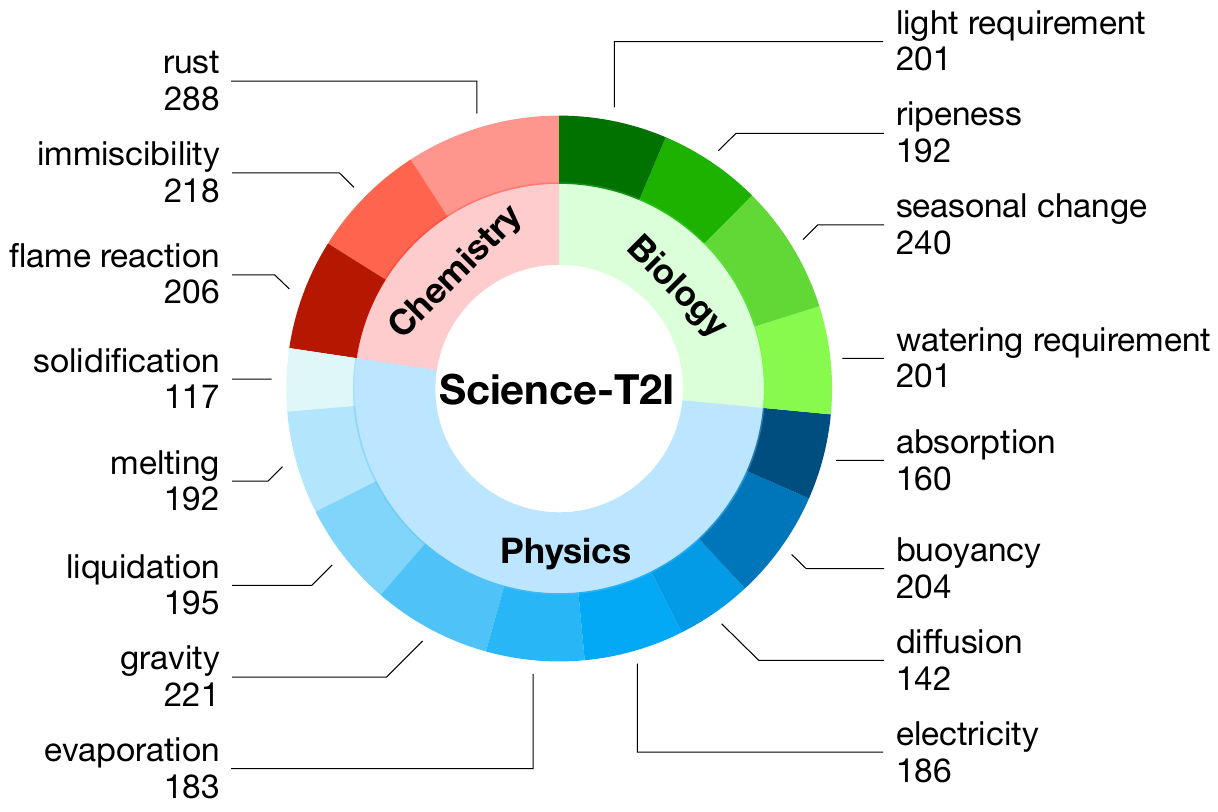}
    \end{minipage}
    \hfill
    \begin{minipage}[b]{0.56\linewidth}
        \includegraphics[width=\linewidth]{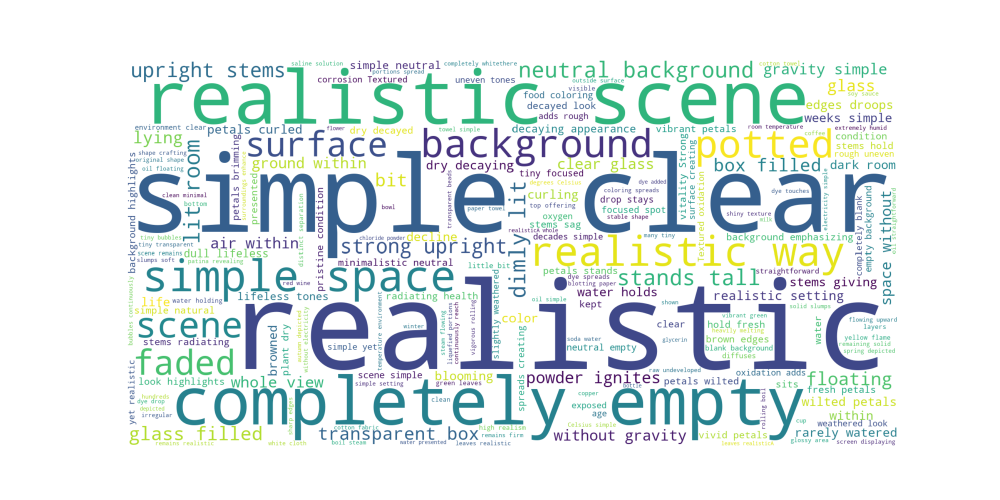}
    \end{minipage}
    \caption{\textbf{Data statistics.} (Left)~\dataset~is organized into three primary scientific fields: Chemistry, Biology, and Physics. Each field is divided into specific categories, with the numbers indicating the volume of implicit prompts collected for each category. (Right) Word cloud of structured prompts in~\dataset.}
    \vspace{-6pt}
    \label{fig:data_stat}
\end{figure}

\section{\dataset: Bridging Visual Imagination and Scientific Realism}
\label{sec:dataset}

As established above, current image generation models produce visually plausible but scientifically incorrect outputs when prompted with descriptions that require physical reasoning. We identify two root causes: (1) existing training data rarely pairs scientific concepts with their correct visual manifestations, and (2) standard evaluation protocols do not test whether a model understands the science behind a prompt or merely matches its surface-level descriptions. To address both issues, we introduce~\dataset, a dataset that challenges models to perform implicit reasoning rather than textual rendering. Unlike conventional datasets that focus on textual descriptions~\cite{5206848, Krizhevsky09learningmultiple, lin2015microsoftcococommonobjects} or preferences~\cite{kirstain2023pickapicopendatasetuser, xu2023imagereward, wu2023human},~\dataset~requires models to infer visual outcomes from prompts grounded in scientific principles.

\paragraph{Task overview.}
As illustrated in Figure~\ref{fig:data_stat},~\dataset~consists of 16 tasks spanning physics, chemistry, and biology. Each task requires the model to infer or visualize a concept not explicitly stated in the prompt but rooted in an underlying scientific principle. These tasks draw on existing research such as PhyBench~\cite{meng2024phybench} and Commonsense-T2I~\cite{fu2024commonsenset2ichallengetexttoimagegeneration}, and extend them with new phenomena developed for this study. Each task satisfies two design objectives:
\begin{itemize}
    \item \textbf{Rewriting Capability.} Prompts allow flexible rephrasing while preserving the same visual meaning. For example, \textit{an unripe apple} can be rephrased as \textit{a green apple}, conveying an identical visual concept. This property enables the construction of explicit and superficial prompt variants from a single implicit prompt, as described below. Further explanations and examples are provided in Appendix~\ref{sec:supp_rewrite_cap}.

    \item \textbf{Scientific Knowledge Integration.} Tasks are grounded in established scientific principles, providing a clear and consistent framework. Compared to commonsense knowledge, which can vary culturally or contextually, scientific principles offer unambiguous ground truth for evaluating correctness.
\end{itemize}
Detailed descriptions of all 16 tasks are provided in Appendix~\ref{sec:supp_tasks}. Beyond the classification by scientific discipline, we observe that the tasks naturally fall into two categories that reveal distinct reasoning demands:
\begin{itemize}
    \item \textbf{Subject-oriented Tasks~(ST)} require reasoning about how inherent differences between subjects lead to varying visual features under identical conditions. For example, different metals produce different flame colors.

    \item \textbf{Condition-oriented Tasks~(CT)} focus on how a single condition affects the visual appearance of various subjects. Here, scientific reasoning centers on the applied condition rather than the subject's individual properties. For example, all objects float in the absence of gravity.
\end{itemize}
As we show in Section~\ref{sec:rw_result}, this distinction proves analytically useful: nearly all failure cases of~\model~concentrate in ST, where subject-specific knowledge is required rather than generalizable visual patterns. The full classification of each task is provided in Appendix~\ref{sec:supp_obs}.

\paragraph{Prompt design.}
A central design choice in~\dataset~is the three-tier prompt structure, which disentangles a model's scientific reasoning ability from its compositional rendering ability. Prior work~\cite{meng2024phybench} observes that models often ignore implicit scientific cues and attend only to surface-level text, motivating us to introduce a third prompt type that captures this failure mode. For each task, we construct a tuple of three prompts:
\begin{itemize}
    \item \textbf{Implicit Prompt~(IP)} contains terms that imply certain visual characteristics requiring interpretative reasoning based on scientific knowledge. For example, ``an unripe apple'' suggests greenness without explicitly stating it.
    \item \textbf{Explicit Prompt~(EP)} reformulates the implicit prompt into a clear, descriptive statement that directly conveys the intended visual outcome. For instance, ``a green apple'' makes the expected appearance explicit.
    \item \textbf{Superficial Prompt~(SP)} provides a plausible but scientifically incorrect interpretation of the implicit prompt, focusing only on surface-level associations. For example, ``a red apple'' interprets ``unripe'' based on the default visual prototype rather than the scientific implication.
\end{itemize}
Together, the three prompt types serve complementary roles: the IP tests implicit reasoning, the EP establishes an upper bound on what the model can render when given direct instructions, and the SP provides a hard negative for preference-based training. Each training tuple in~\dataset~pairs these prompts with corresponding explicit and superficial images, forming the adversarial format required for reward modeling (Section~\ref{sec:phy_train}).

\begin{figure*}[t]
    \centering
    \includegraphics[width=0.98\linewidth]{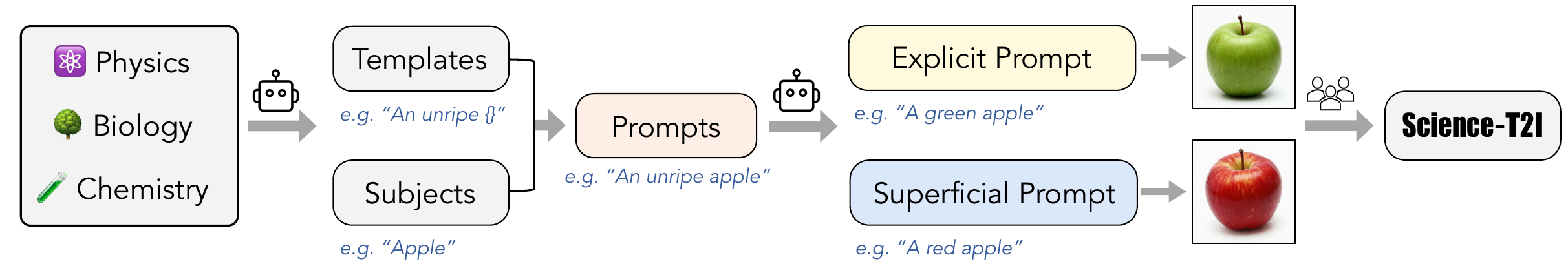}
    \caption{\textbf{Data curation pipeline}. For each task, \texttt{GPT-4o}~\cite{GPT} first generates structured templates that capture the scientific principles while allowing for variability in objects or substances. These templates are used to create implicit prompts, which \texttt{GPT-4o}~\cite{GPT} then expands into explicit and superficial prompts, ultimately guiding the synthesis of corresponding explicit and superficial images.}
    \label{fig:data_pipe}
    \vspace{-9pt}
\end{figure*}

\paragraph{Data curation.}
We leverage \texttt{GPT-4o}~\cite{GPT} to generate structured templates and corresponding prompts for each task. As illustrated in Figure~\ref{fig:data_pipe}, the pipeline first produces implicit prompts from task-specific templates, then expands each implicit prompt into its explicit and superficial counterparts. These prompts are used to generate image pairs via image generation models. To ensure quality, all generated data undergoes manual verification by domain experts, who cross-reference each tuple against established scientific knowledge. Further details are provided in Appendix~\ref{sec:supp_dataset_detail}.

\section{Benchmarking Scientific Image Synthesis}
\label{sec:benchmark}

Having introduced the~\dataset~training set and its prompt structure, we now turn to evaluation. We construct a dedicated test set, define a two-tiered grading framework, and use it to benchmark recent image generation models. The results expose a systematic gap between compositional rendering and scientific reasoning.

\paragraph{Test splits.}
We construct two manually annotated test sets covering the same scientific domains as the training data but containing no overlapping samples. Each tuple is reviewed iteratively by domain experts who cross-reference their annotations with authoritative scientific sources until unanimous consensus is reached. To evaluate generation quality under varying visual complexity, we split the benchmark into two subsets:
\begin{itemize}
    \item \textbf{\dataset-S (Simple)}, containing 227 tuples. This split uses minimalist compositions with simple backgrounds. By isolating the scientific subject from environmental distractions, it provides a controlled test of whether the model can generate the correct scientific visual features.
    \item \textbf{\dataset-C (Complex)}, containing 227 tuples. This split embeds the same scientific tasks within diverse real-world scenes, adding contextual elements such as ``in a bedroom'' or ``on the street.'' It tests a more challenging setting: whether the model can maintain scientific accuracy when the background is visually complex and potentially distracting. Strong performance on~\dataset-C indicates generalization beyond the training distribution.
\end{itemize}
For brevity, we refer to the union of both splits as the~\dataset~test set below.

\begin{table*}[t]
  \caption{\textbf{Current image generation models struggle with implicit scientific reasoning across all domains.} We evaluate 18 models on~\dataset~using implicit prompts and report the normalized Reality Score evaluated by Qwen3.5-27B~\cite{qwen3.5} for Physics, Chemistry, and Biology. Even the strongest model scores below 50, and Biology consistently lags behind the other two domains. \best{Bold} and \secondbest{underlined} denote the best and second-best results.}
  \label{tab:alignment}
  \small
  \centering
  \setlength{\tabcolsep}{6pt}
  \begin{tabular}{l|C{1.3cm}C{1.3cm}C{1.3cm}C{1.3cm}}
    \toprule
    \multirow{2}{*}{Model} & \multicolumn{4}{c}{\dataset} \\
    \cmidrule(lr){2-5}
    & {Physics} & {Chemistry} & {Biology} & {Overall} \\
    \midrule
    SDXL~\cite{podell2023sdxlimprovinglatentdiffusion}                  & $16.11$ & $20.92$ & $25.56$ & $19.60$\\
    FLUX.1[schnell]~\cite{FLUX}       & $23.19$ & $38.30$ & $13.33$ & $23.72$\\
    LongCat-Image~\cite{longcat}         & $21.94$ & $39.01$ & $15.83$ & $23.86$\\
    GLM-Image~\cite{glmimage}             & $22.22$ & $41.84$ & $14.44$ & $24.23$\\
    SANA~\cite{xie2024sana}                  & $23.89$ & $46.45$ & $14.72$ & $26.14$\\
    Lumina-Image-2~\cite{qin2025luminaimage20unifiedefficient}        & $20.28$ & $47.87$ & $22.22$ & $26.51$\\
    SD 3.5 Medium~\cite{sd3}         & $23.06$ & $39.01$ & $24.44$ & $26.73$\\
    Z-Image~\cite{imageteam2025zimageefficientimagegeneration}               & $26.53$ & $32.98$ & $22.22$ & $26.73$\\
    FLUX.1[dev]~\cite{FLUX}           & $22.64$ & $50.00$ & $17.22$ & $26.87$\\
    SD 3.5 Large~\cite{sd3}          & $27.64$ & $44.68$ & $18.33$ & $28.71$\\
    Bagel~\cite{deng2025emergingpropertiesunifiedmultimodal}                 & $25.28$ & $45.75$ & $23.33$ & $29.00$\\
    FLUX.2[klein-4B-Base]~\cite{flux-2-2025} & $27.22$ & $43.97$ & $21.11$ & $29.08$\\
    Z-Image-Turbo~\cite{imageteam2025zimageefficientimagegeneration}         & $26.81$ & $36.53$ & \secondbest{28.33} & $29.22$\\
    FLUX.2[klein-4B]~\cite{flux-2-2025}      & $28.75$ & $50.36$ & $19.72$ & $30.84$\\
    FLUX.2[klein-9B-Base]~\cite{flux-2-2025} & $30.28$ & $45.75$ & $23.06$ & $31.57$\\
    FLUX.2[klein-9B]~\cite{flux-2-2025}      & $29.44$ & \best{57.45} & $19.44$ & $32.60$\\
    Qwen-Image~\cite{wu2025qwenimagetechnicalreport}            & \secondbest{34.58} & $46.45$ & $23.06$ & \secondbest{33.99}\\
    FLUX.2[dev]~\cite{flux-2-2025}           & \best{53.19} & \secondbest{53.55} & \best{32.50} & \best{47.80}\\
    \bottomrule
  \end{tabular}
  \vspace{-6pt}
\end{table*}

\paragraph{Grading criteria.}
We leverage the vision and reasoning capabilities of LMMs to assess generated images. To ensure reliable and fine-grained evaluation, we develop per-tuple grading criteria inspired by PhyBench~\cite{meng2024phybench}. Each tuple is scored along two complementary aspects:
\begin{itemize}
    \item \textbf{Scene Score~(SS)} measures whether all descriptive visual content specified in the prompt is faithfully present in the generated image. The full score is 2.
    \item \textbf{Reality Score~(RS)} assesses whether the implicit visual elements, derived from the underlying scientific principles, are accurately depicted in the image. The full score is 3.
\end{itemize}
We present representative examples of the grading criteria and corresponding prompts in Figure~\ref{fig:sample_exp}.

\paragraph{Scoring protocol.}
For each tuple in the~\dataset~test set, we use the implicit prompt to generate two images with each model. We then employ Qwen3.5-27B~\cite{qwen3.5} as the evaluator following the instruction template detailed in Figure~\ref{fig:user_inst}. For every generated image, Qwen3.5-27B~\cite{qwen3.5} receives the image alongside the implicit prompt and its grading criteria, and produces its SS and RS scores. Notably, within this two-tiered framework, assessing scientific correctness is meaningful only when the main subject is correctly depicted. We therefore set RS to zero for any image that does not achieve a full SS. To obtain the final metric, we normalize the RS to the range of $[0, 100]$.

\paragraph{Models.}
We evaluate 18 image generation models spanning diverse architectures and scales. These include the Stable Diffusion series~\cite{podell2023sdxlimprovinglatentdiffusion,sd3}, the FLUX series~\cite{FLUX}, and recent models such as GLM-Image~\cite{glmimage}, Qwen-Image~\cite{wu2025qwenimagetechnicalreport}, Z-Image~\cite{imageteam2025zimageefficientimagegeneration}, Bagel~\cite{deng2025emergingpropertiesunifiedmultimodal}, SANA~\cite{xie2024sana}, Lumina-Image-2~\cite{qin2025luminaimage20unifiedefficient}, and LongCat-Image~\cite{longcat}. This selection covers both open-source and proprietary systems released between 2022 and 2025, providing a representative snapshot of current capabilities. The full results are presented in Table~\ref{tab:alignment}.

\paragraph{Current models lack scientific image synthesis capabilities.}
As shown in Table~\ref{tab:alignment}, all 18 models exhibit consistently low scores on~\dataset~under implicit prompts. Even the best-performing model, FLUX.2[dev]~\cite{flux-2-2025}, achieves only 47.80 out of 100, and the majority of models cluster between 20 and 35. Among the three domains, Biology poses the greatest challenge: no model exceeds 33\%, suggesting that biological reasoning remains particularly underrepresented in current training data. A telling example is Z-Image~\cite{imageteam2025zimageefficientimagegeneration}, a model widely recognized for its photorealistic quality and strong compositional capabilities. Despite being a recent and highly competitive model, it scores only 26.73 on~\dataset, lower than the significantly earlier FLUX.1[dev]~\cite{FLUX} at 26.87. This observation indicates that advances in visual fidelity and scene composition do not translate into scientific reasoning ability, and that current training pipelines systematically lack data targeting this capability.

\begin{figure*}[t]
    \centering
    \includegraphics[width=0.98\linewidth]{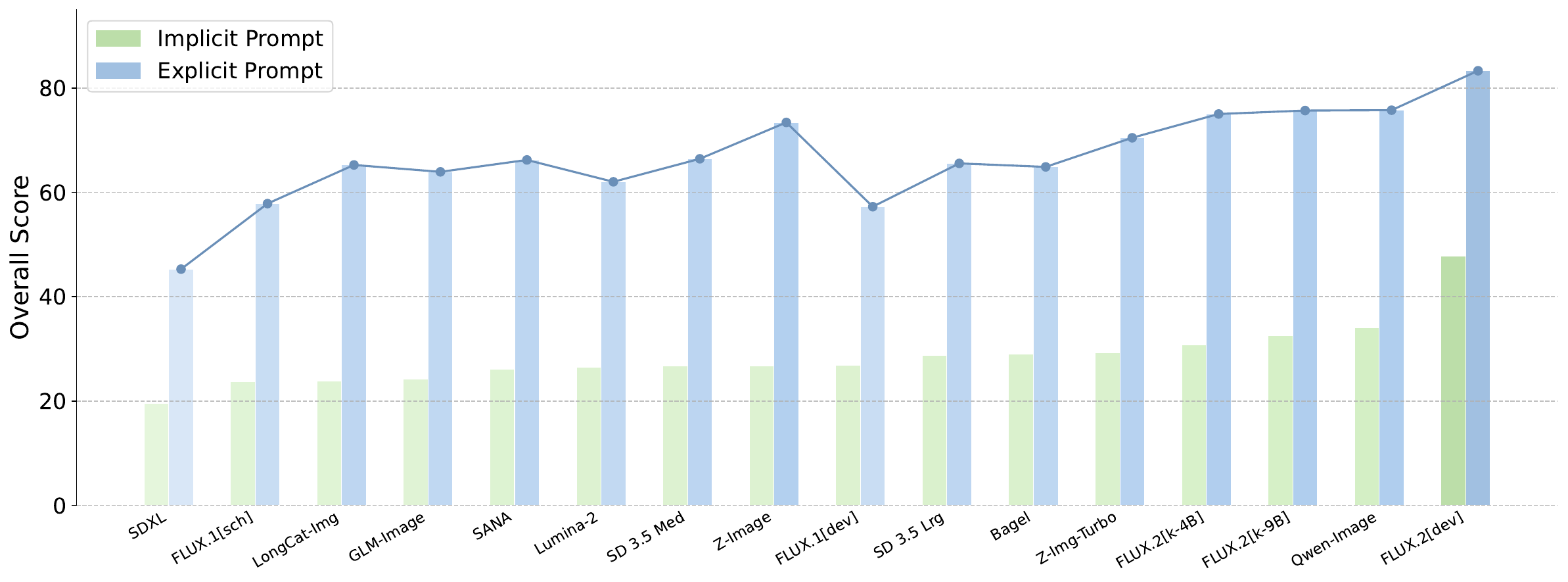}
    \caption{\textbf{Explicit prompts close the gap that implicit prompts expose.} All models score significantly higher under explicit prompts (blue) than implicit prompts (green), with an average gap of roughly 35 points. The trend line highlights that even as overall capability increases, the composition--reasoning disconnect persists across all models.}
    \label{fig:ep_vs_ip}
\end{figure*}

\paragraph{Explicit prompts reveal a composition--reasoning disconnect.}
The results above show that models perform poorly under implicit prompts, but they do not reveal whether the bottleneck lies in visual rendering or in scientific reasoning. To disentangle these two factors, we replace each implicit prompt with its corresponding explicit prompt and re-evaluate all models using Qwen3.5-27B~\cite{qwen3.5}. The results, shown in Figure~\ref{fig:ep_vs_ip}, reveal a striking gap: across all 18 models, explicit prompts yield scores roughly 35 points higher than implicit prompts on average. This demonstrates that current models can faithfully render scientifically correct scenes when explicitly told what to depict, yet fail to infer the same visual outcome from implicit scientific cues. The contrast is especially pronounced for Z-Image~\cite{imageteam2025zimageefficientimagegeneration}, which achieves 73.42 under explicit prompts but only 26.73 under implicit prompts, a gap of nearly 47 points. This finding clarifies that the bottleneck is not in visual rendering but in scientific reasoning: models excel at translating direct descriptions into images, but lack the capacity to reason from scientific principles to their visual consequences.

\section{\model: A Reward Model Grounded in Reality}

CLIP~\cite{radford2021learningtransferablevisualmodels} aligns textual and visual data effectively for general purposes, but we find that it struggles with implicit scientific prompts: it tends to embed implicit prompts closer to their superficial counterparts than their explicit counterparts, because surface-level associations dominate over scientific implications. To address this, we introduce~\model, a reward model fine-tuned on the~\dataset~training set that extends CLIP to assess whether a generated image reflects the scientific principles implied by a prompt. We first define the reward formulation (\S\ref{sec:reward_model}) and then describe the two training objectives that jointly optimize~\model~(\S\ref{sec:phy_train}).

\subsection{Reward Formulation}
\label{sec:reward_model}
\model~builds on the CLIP architecture~\cite{radford2021learningtransferablevisualmodels}, encoding a text prompt $x$ and an image $y$ into a shared high-dimensional space using separate transformer encoders~\cite{vaswani2023attentionneed}, $E_{\text{txt}}$ and $E_{\text{img}}$. The reward is the cosine similarity between the two representations, scaled by a learnable temperature $T$:
\begin{equation}
    r(y, x) = T \cdot \frac{E_{\text{txt}}(x) \cdot E_{\text{img}}(y)}{\| E_{\text{txt}}(x) \| \, \| E_{\text{img}}(y) \|}.
\end{equation}

\subsection{Training Objective}
\label{sec:phy_train}
We fine-tune CLIP-H~\cite{ilharco_gabriel_2021_5143773} on the~\dataset~training set with both encoders learnable. Each training instance is a tuple $(x_i, x_e, x_s, y_e, y_s)$: an implicit prompt $x_i$, its explicit and superficial counterparts $x_e$ and $x_s$, and the corresponding explicit and superficial images $y_e$ and $y_s$. The training objective combines two complementary losses targeting the text and image encoders, respectively.

\paragraph{Preference formulation.} Following preference modeling in language~\cite{stiennon2022learningsummarizehumanfeedback, ouyang2022traininglanguagemodelsfollow}, the predicted preference $\hat{p}_{\text{img}}(x_a \succ x_b; y)$ for prompt $x_a$ over prompt $x_b$ for a given image $y$ is calculated as:
\begin{equation}
    \hat{p}_{\text{img}}(x_a \succ x_b; y) = \frac{\exp(r(y, x_a))}{\exp(r(y, x_b)) + \exp(r(y, x_a))}
\end{equation}
Similarly, for a given prompt $x$, the predicted preference $\hat{p}_{\text{txt}}(y_a \succ y_b; x)$ for image $y_a$ over image $y_b$ is given by:
\begin{equation}
    \hat{p}_{\text{txt}}(y_a \succ y_b; x) = \frac{\exp(r(y_a, x))}{\exp(r(y_a, x)) + \exp(r(y_b, x))}
\end{equation}

\paragraph{Implicit prompt alignment (IPA).} The core challenge is that pretrained CLIP embeds implicit prompts in a manner similar to their superficial counterparts rather than their explicit counterparts, because surface-level co-occurrence patterns dominate over scientific semantics. IPA corrects this by minimizing the KL divergence between the target preference $p_{\text{txt}} = [1, 0]$ and the predicted preference $\hat{p}_{\text{txt}} = \left[\hat{p}_{\text{txt}}(y_e \succ y_s; x_i),\, \hat{p}_{\text{txt}}(y_s \succ y_e; x_i)\right]$, effectively steering the implicit prompt embedding toward the explicit image:
\begin{equation}
    \mathcal{L}_{\text{IPA}} = \sum_{j=1}^{2} p_{\text{txt}_j} \left( \log p_{\text{txt}_j} - \log \hat{p}_{\text{txt}_j} \right)
\end{equation}

\paragraph{Image encoder enhancement (IEE).}
IPA operates exclusively on the text encoder's representation of the implicit prompt. However, many scientific phenomena manifest as fine-grained visual details (e.g., subtle color differences or layering patterns) that the pretrained image encoder may not attend to. To strengthen the encoder's sensitivity to these details, we introduce a complementary loss on the image side:
\begin{equation}
    \mathcal{L}_{\text{IEE}} = \mathcal{L}_{\text{img}}^{+} + \mathcal{L}_{\text{img}}^{-},
\end{equation}
where $\mathcal{L}_{\text{img}}^{+}$ and $\mathcal{L}_{\text{img}}^{-}$ correspond to the losses associated with explicit and superficial image preferences, respectively. The explicit image loss $\mathcal{L}_{\text{img}}^{+}$ is defined as:
\begin{equation}
    \mathcal{L}_{\text{img}}^{+} = \sum_{j=1}^2 p_{\text{img}_j}^{+} \left( \log p_{\text{img}_j}^{+} - \log \hat{p}_{\text{img}_j}^{+} \right),
\end{equation}
with $p_{\text{img}}^{+} = [1, 0]$ and $\hat{p}_{\text{img}}^{+} = [\hat{p}_{\text{img}}(x_e \succ x_s; y_e),\, \hat{p}_{\text{img}}(x_s \succ x_e; y_e)]$.
The superficial image loss is defined symmetrically:
\begin{equation}
    \mathcal{L}_{\text{img}}^{-} = \sum_{j=1}^2 p_{\text{img}_j}^{-} \left( \log p_{\text{img}_j}^{-} - \log \hat{p}_{\text{img}_j}^{-} \right),
\end{equation}
with $p_{\text{img}}^{-} = [0, 1]$ and $\hat{p}_{\text{img}}^{-} = [\hat{p}_{\text{img}}(x_e \succ x_s; y_s),\, \hat{p}_{\text{img}}(x_s \succ x_e; y_s)]$.
The overall loss combines both objectives:
\begin{equation}
    \mathcal{L} = \mathcal{L}_{\text{IPA}} + \lambda \mathcal{L}_{\text{IEE}},
\end{equation}
where $\lambda$ controls the relative weight of image encoder enhancement. We ablate $\lambda$ in Section~\ref{sec:rw_result}.

\subsection{Experimental Setup}

\paragraph{Training and evaluation.} We fine-tune CLIP-H~\cite{ilharco_gabriel_2021_5143773} on the~\dataset~training set with both text and image encoders learnable. Training completes within one hour on 8 A6000 GPUs. We evaluate on both~\dataset-S and~\dataset-C. For detailed configurations, see Appendix~\ref{sec:supp_train_sciscore}.

\paragraph{Baselines.} We compare against three categories of baselines: (1) VLMs including CLIP-H~\cite{ilharco_gabriel_2021_5143773}, BLIPScore~\cite{li2022blipbootstrappinglanguageimagepretraining, li2023blip2bootstrappinglanguageimagepretraining}, and SigLIP~\cite{zhai2023sigmoidlosslanguageimage}; (2) LMMs including LLaVA-OV~\cite{li2024llavaonevisioneasyvisualtask}, Qwen2-VL~\cite{wang2024qwen2vlenhancingvisionlanguagemodels}, InternVL~\cite{chen2024internvlscalingvisionfoundation}, and \texttt{GPT-4o-mini}~\cite{GPT}, with and without CoT~\cite{wei2023chainofthoughtpromptingelicitsreasoning}; and (3) human evaluations by 10 experts with science or engineering degrees. Details are provided in Appendix~\ref{sec:supp_baseline_sciscore}; results are summarized in Table~\ref{tab:pref_acc}.

\begin{table*}[t]
  \caption{\textbf{\model~surpasses all VLMs, LMMs, and human evaluators.} Accuracy (\%) on the two-choice selection task across Physics, Chemistry, and Biology. \colorbox{TabGreen}{Green} highlights the best VLM; \colorbox{TabBlue}{blue} highlights the best LMM.}
  \label{tab:pref_acc}
  \centering
  \resizebox{0.95\linewidth}{!}{
  \begin{tabular}{l|cccc|cccc}
    \toprule
         \multirow{2}{*}{Model} & \multicolumn{4}{c|}{\dataset-S} & \multicolumn{4}{c}{\dataset-C} \\
    \cmidrule(lr){2-5} \cmidrule(lr){6-9}
    & {Physics} & {Chemistry} & {Biology} & {Avg.} & {Physics} & {Chemistry} & {Biology} & {Avg.} \\
    \midrule
    Human Eval & $87.67$ & $75.85$ & $95.29$ & $87.01$ & $84.71$ & $85.40$ & $89.14$ & $86.02$ \\
    Random Guess & $50.00$ & $50.00$ & $50.00$ & $50.00$ & $50.00$ & $50.00$ & $50.00$ & $50.00$ \\
    \midrule
    CLIP-H~\cite{ilharco_gabriel_2021_5143773} & $55.08$ & $52.38$ & $55.88$ & $54.69$ & $56.56$ & $44.44$ & \cellcolor{TabGreen}$76.67$ & $59.47$ \\
    BLIPScore~\cite{li2022blipbootstrappinglanguageimagepretraining} & $50.35$ & $43.08$ & \cellcolor{TabGreen}$59.86$ & $55.00$ & $49.78$ & \cellcolor{TabGreen}$60.00$ & $58.33$ & $51.54$ \\
    SigLIP ViT-SO-14~\cite{zhai2023sigmoidlosslanguageimage} & \cellcolor{TabGreen}$59.63$ & \cellcolor{TabGreen}$53.17$ & $55.94$ & \cellcolor{TabGreen}$57.23$ & \cellcolor{TabGreen}$61.48$ & $51.11$ & $70.00$ & \cellcolor{TabGreen}$61.67$ \\
    \midrule
    Qwen2-VL-7B~\cite{wang2024qwen2vlenhancingvisionlanguagemodels} & $60.03$ & $67.01$ & $68.82$ & $63.79$ & $66.80$ & $50.00$ & $90.83$ & $69.82$ \\
    LLaVA-OV-7B~\cite{li2024llavaonevisioneasyvisualtask} & \cellcolor{TabBlue}$68.22$ & $57.82$ & $64.71$ & $65.05$ & \cellcolor{TabBlue}$74.59$ & $50.00$ & $76.67$ & $70.26$ \\
    InternVL2.5-8B~\cite{chen2024internvlscalingvisionfoundation} & $67.80$ & $62.24$ & $84.41$ & $70.79$ & $73.77$ & $65.56$ & $85.83$ & $75.33$ \\
    GPT-4o mini~\cite{GPT} & $61.97$ & $73.81$ & $86.76$ & $70.83$ & $69.29$ & \cellcolor{TabBlue}$70.00$ & $90.00$ & $74.78$ \\
    GPT-4o mini$+$~CoT~\cite{wei2023chainofthoughtpromptingelicitsreasoning} & $67.04$ & \cellcolor{TabBlue}$76.87$ & \cellcolor{TabBlue}$90.00$ & \cellcolor{TabBlue}$74.97$ & $72.44$ & \cellcolor{TabBlue}$70.00$ & \cellcolor{TabBlue}$92.50$ & \cellcolor{TabBlue}$77.16$ \\
    \midrule
    \model~(ours) & \textbf{94.92} & \textbf{80.95} & \textbf{100.0} & \textbf{93.14} & \textbf{86.89} & \textbf{91.11} & \textbf{100.0} & \textbf{91.19} \\
    \bottomrule
  \end{tabular}
  }
\end{table*}

\subsection{\model~Outperforms VLMs, LMMs, and Human Evaluators}
\label{sec:rw_result}
\paragraph{VLMs and LMMs fail to distinguish scientifically accurate images.}
As shown in Table~\ref{tab:pref_acc}, all three VLMs achieve near-random accuracy (54--61\%) across both test splits, indicating that general-purpose vision-language alignment provides little signal for scientific reasoning. LMMs fare better but remain far below human performance: even the strongest baseline, \texttt{GPT-4o-mini} with CoT~\cite{wei2023chainofthoughtpromptingelicitsreasoning}, reaches only 74.97 on~\dataset-S and 77.16 on~\dataset-C, compared to human scores of 87.01 and 86.02. Notably, CoT prompting yields only marginal improvements over the standard \texttt{GPT-4o-mini}, suggesting that the bottleneck lies in visual perception rather than reasoning chain construction. A detailed analysis is provided in Appendix~\ref{sec:supp_analysis_sciscore}.

\paragraph{\model~generalizes to complex scenes and surpasses human evaluators.}
\model~achieves 93.14 on~\dataset-S and 91.19 on~\dataset-C, surpassing human evaluators (87.01 and 86.02) by roughly 6 and 5 points, respectively. The strong performance on~\dataset-C is particularly notable, as this split embeds scientific reasoning tasks within diverse environmental contexts that were not present during training. This suggests that~\model~has learned to focus on the scientifically relevant regions of an image while disregarding environmental distractions.

\paragraph{Subject-oriented tasks remain the primary challenge.}
We further analyze~\model's performance by the ST/CT classification introduced in Section~\ref{sec:dataset}. As shown in Figure~\ref{fig:st_ct}, nearly all failure cases concentrate in subject-oriented tasks. This is expected: condition-oriented tasks rely on generalizable visual features (e.g., the absence of gravity implies floating objects), whereas subject-oriented tasks demand subject-specific knowledge (e.g., which metals produce which flame colors). For novel or unseen subjects, the model lacks sufficient prior exposure to the subject's distinctive visual attributes.

\paragraph{$\lambda = 0.25$ balances fine-grained detection with prompt alignment.}
We ablate the IEE weight $\lambda$ in Table~\ref{tab:abla_lambda}. Setting $\lambda = 0$ disables IEE entirely, yielding optimal performance on~\dataset-S but degraded accuracy on~\dataset-C, where fine-grained visual discrimination is more critical due to complex backgrounds. Larger values ($\lambda \geq 0.5$) shift too much focus to the image encoder at the expense of prompt alignment. We find that $\lambda = 0.25$ achieves the best balance, matching the top score on both splits simultaneously. Qualitative analysis is provided in Appendix~\ref{sec:supp_abla_sciscore}.

\clearpage

\begin{table}[t]
  \centering
  \begin{minipage}{0.42\linewidth}
    \centering
    \caption{\textbf{$\lambda = 0.25$ achieves the best balance between IPA and IEE.} \textbf{Bold} indicates best performance per column.}
    \resizebox{\linewidth}{!}{
    \begin{tabular}{c|cc}
      \toprule
      $\lambda$ & \dataset-S & \dataset-C \\
      \midrule
      $0$ & \textbf{93.14} & $88.99$ \\
      $0.1$ & $92.85$ & $90.75$ \\
      $0.5$ & $92.85$ & \textbf{91.19}\\
      $0.75$ & \textbf{93.14} & $88.99$\\
      \midrule
      $0.25$ & \textbf{93.14} & \textbf{91.19}\\
      \bottomrule
    \end{tabular}
    }
    \label{tab:abla_lambda}
  \end{minipage}
  \hfill
  \begin{minipage}{0.55\linewidth}
    \centering
    \includegraphics[width=0.98\linewidth]{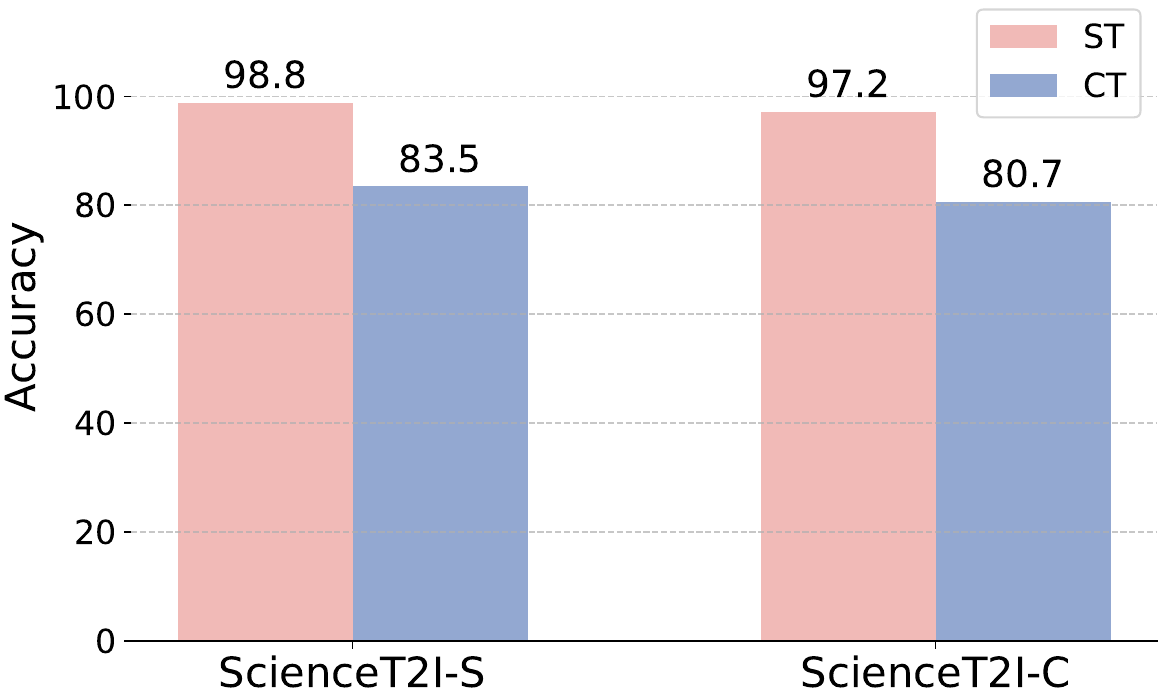}
    \captionof{figure}{\textbf{Nearly all~\model~failures concentrate in subject-oriented tasks.} Condition-oriented tasks involve generalizable visual patterns and are substantially easier.}
    \label{fig:st_ct}
  \end{minipage}
  \vspace{-9pt}
\end{table}

\section{Aligning Generative Models with Scientific Principles}

We now combine the~\dataset~training set with~\model~as a reward signal to close the reasoning gap identified in Section~\ref{sec:benchmark}. Our framework proceeds in two stages: SFT exposes the model to scientific visual phenomena it has never encountered during pretraining, and OFT then optimizes for the implicit reasoning ability measured by~\model.

\subsection{SFT Provides the Scientific Knowledge Base}

Current post-training algorithms for diffusion models, such as those based on PPO~\cite{black2024trainingdiffusionmodelsreinforcement, fan2023dpokreinforcementlearningfinetuning} and DPO~\cite{yang2024usinghumanfeedbackfinetune, wallace2023diffusionmodelalignmentusing}, optimize within the distribution of the pre-trained model. This constraint is acceptable for aesthetic preferences, where the desired outputs already lie within the model's distribution. However, scientific reasoning presents a fundamentally different challenge: pre-trained models have never been exposed to images depicting scientific phenomena, so no amount of preference optimization over existing outputs can teach them what these phenomena look like. The model must first learn what scientifically correct images are before it can be steered toward generating them.

We therefore begin with supervised fine-tuning on the~\dataset~training set to provide this missing foundation. Among the models evaluated in Section~\ref{sec:benchmark}, the FLUX series~\cite{FLUX} consistently achieves the strongest text-image alignment and produces the most realistic outputs. We adopt FLUX.1[dev]~\cite{FLUX} as our base model. Since FLUX employs the flow matching~\cite{lipman2023flowmatchinggenerativemodeling} framework, the SFT objective is:
\begin{equation}
    L_{SFT} = \mathbb{E}_{t, p_t(z|\epsilon), p(\epsilon)} \left\| v_\theta(z, t) - u_t(z | \epsilon) \right\|_2^2
\end{equation}
We adopt the same notation as~\cite{sd3}; further details are provided in Appendix~\ref{sec:supp_two_stage_train_detail}.

\subsection{Online Fine-tuning with \model~as Reward Signal}
\begin{figure*}[t]
    \centering
    \includegraphics[width=0.98\linewidth]{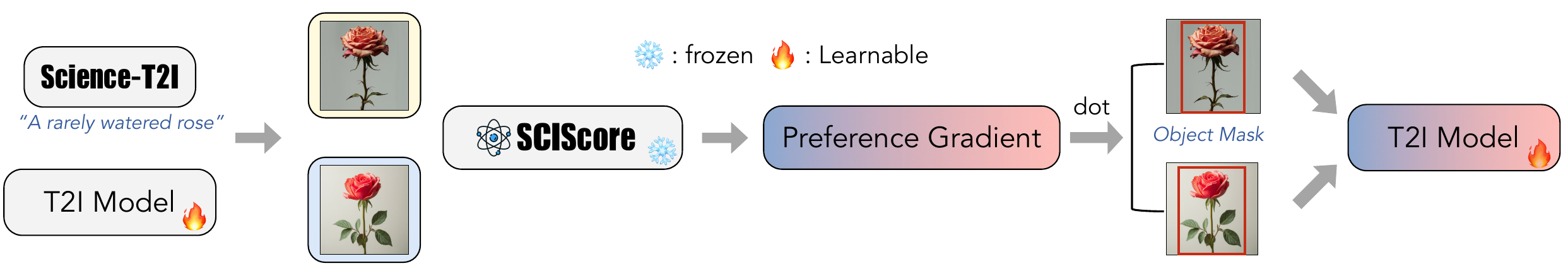}
    \caption{\textbf{Online fine-tuning uses~\model~and subject-based masking to align generation with scientific principles.} For each prompt, two images are generated and scored by~\model~to determine preference. GroundingDINO~\cite{liu2024groundingdinomarryingdino} extracts subject masks from each image, restricting gradient propagation to the relevant regions.}
    \label{fig:tuning}
    \vspace{-9pt}
\end{figure*}

SFT teaches the model \textit{what} scientific images look like by training on explicit image pairs. However, it does not directly optimize for the ability to infer the correct visual outcome from an implicit prompt, which is the core capability measured by~\model. To bridge this remaining gap, we apply online fine-tuning with~\model~as the reward signal, as illustrated in Figure~\ref{fig:tuning}. Following DDPO~\cite{black2024trainingdiffusionmodelsreinforcement}, we formulate the denoising process as a multi-step MDP:
\begin{equation}
    s_t \triangleq (c, t, x_{1-t}),\ \pi_\theta(a_t \mid s_t) \triangleq p_\theta (x_{1-\Delta t-t} \mid c, t, x_{1-t}),\ \rho_0 (s_0) \triangleq (p(c), \delta_0, \mathcal{N}(0, I))
\end{equation}
\begin{equation}
    a_t \triangleq x_{1-\Delta t-t},\ P(s_{t+\Delta t} \mid s_t, a_t) \triangleq (\delta_c, \delta_{t+\Delta t}, \delta_{x_{1-t-\Delta t}}),\ r(s_t, a_t) \triangleq \begin{cases}
    r(x_0, c) & \text{if } t = 1 \\
    0 & \text{otherwise}
    \end{cases}
\end{equation}
We follow the notation of DDPO~\cite{black2024trainingdiffusionmodelsreinforcement} with minor adjustments to the timestamp convention. However, flow matching~\cite{lipman2023flowmatchinggenerativemodeling} is typically formulated as an ODE, yielding a deterministic policy that prevents direct computation of the log-probability required by DPO:
\begin{equation}
\label{equ:15}
\pi_\theta(a_t \mid s_t) = \delta\left(x_{1-\Delta t-t} - (x_{1-t} - v_\theta(s_t)\Delta t)\right)
\end{equation}
Following~\cite{domingoenrich2024adjointmatchingfinetuningflow}, we resolve this by interpreting flow matching as an SDE:
\begin{equation}
\mathrm{d}x_t = \left( v_\theta(x_t, t) + \frac{\sigma_t^2}{2\beta_t\eta_t} \lambda_t\right) \mathrm{d}t + \sigma_t \mathrm{d}B_t
\end{equation}
\begin{equation}
\eta_t = \left(\frac{\dot{\alpha}_t}{\alpha_t} \beta_t - \dot{\beta}_t\right), \quad \lambda_t = \left( v(x_t, t) - \frac{\dot{\alpha}_t}{\alpha_t} x_t \right)
\end{equation}
where $B_t$ denotes Brownian motion. Discretizing with the rectified flow used by FLUX~\cite{FLUX} ($\alpha_t = t$, $\beta_t = 1-t$) yields a Gaussian policy:
\begin{equation}
\pi_\theta (a_t \mid s_t) = \mathcal{N}\left(a_t;\mu_\theta(s_t), \sigma_t^2I\right)
\end{equation}
\begin{equation}
\mu_\theta(s_t) = \frac{t\sigma_t^2+2(1-t)}{-2(1-t)}v_\theta(s_t)\Delta t+\frac{2(1-t)+\sigma_t^2\Delta t}{2(1-t)}x_{1-t}
\end{equation}
Setting $\sigma_t = 0$ recovers the deterministic case in Equation~\ref{equ:15}. For the training objective, we adopt DPO~\cite{rafailov2024directpreferenceoptimizationlanguage}, with technical details in Appendix~\ref{sec:supp_two_stage_train_detail}. Given a prompt $c$, we sample two denoising trajectories:
\begin{equation}
    \sigma_w = \{s_0^w, a_0^w, s_{\Delta t}^w, a_{\Delta t}^w, \dots, s_1^w, a_1^w\},\ \sigma_l = \{s_0^l, a_0^l, s_{\Delta t}^l, a_{\Delta t}^l, \dots, s_1^l, a_1^l\}
\end{equation}
Assuming that the reward satisfies $r(s_1^w, a_1^w) > r(s_1^l,a_1^l)$, the training objective is formulated as:
\begin{equation}
    \mathcal{L} = \mathbb{E}\left[\log \rho \left(\beta\log \frac{\pi_\theta(a_k^l | s_k^l)}{\pi_\text{ref}(a_k^l | s_k^l)} - \beta\log \frac{\pi_\theta(a_k^w | s_k^w)}{\pi_\text{ref}(a_k^w | s_k^w)} \right) \right]
\end{equation}

\paragraph{Subject-based masking for stable optimization.} Without masking, the DPO objective treats all spatial regions of the preferred image as equally desirable. In practice, the preferred and rejected images often differ only in the scientifically relevant region, while sharing similar backgrounds and compositions. Optimizing over the entire image introduces noise from irrelevant features and destabilizes training. To address this, we extract the subject from each prompt and use GroundingDINO~\cite{liu2024groundingdinomarryingdino} to localize it within the generated image. Only the content within the detected bounding box contributes to gradient backpropagation. We define the mask as $\mathcal{M}$; the masked objective becomes:

\begin{equation}
    \mathcal{L} = -\mathbb{E}\Bigg[ \log \rho \bigg(\beta \log \frac{\mathcal{M}^{w} \odot \pi_\theta(a_k^w \mid s_k^w)}{\mathcal{M}^{w} \odot \pi_\text{ref}(a_k^w \mid s_k^w)} \notag - \beta \log \frac{\mathcal{M}^{l} \odot \pi_\theta(a_k^l \mid s_k^l)}{\mathcal{M}^{l} \odot \pi_\text{ref}(a_k^l \mid s_k^l)} \bigg) \Bigg].
\end{equation}

\subsection{Training and Evaluation Setup}
\paragraph{Training.}
We first fine-tune FLUX.1[dev]~\cite{FLUX} on~\dataset~using SFT with LoRA~\cite{hu2021loralowrankadaptationlarge} for 2,000 steps, producing LoRA weights for the subsequent OFT stage. For OFT, we randomly select 300 implicit prompts from~\dataset~as the training set. During each epoch, 32 prompts are sampled; for each prompt, two images are generated and scored by~\model. Subject masks are extracted using GroundingDINO~\cite{liu2024groundingdinomarryingdino}. The model is then fine-tuned for approximately 100 steps. Detailed configurations are provided in Appendix~\ref{sec:supp_ft_train_setting}.

\paragraph{Evaluation.}
We extract all implicit prompts from~\dataset-S and~\dataset-C to form two evaluation sets. For each prompt, we generate five images and report the average~\model~to ensure robust results.

\paragraph{Relative improvement metric.}
Raw~\model~scores are useful for comparing models but do not reveal how close a fine-tuned model is to its own ceiling. We observe that~\model~under explicit prompts consistently surpasses that under implicit prompts, providing a natural upper bound: the best the model could achieve if it could perfectly reason from implicit cues. To quantify progress toward this ceiling, we define the \textbf{Relative Improvement} (RI) metric. Let $\text{SciScore}_B^{IP}$ and $\text{SciScore}_B^{EP}$ denote the base model's scores under implicit and explicit prompts, and $\text{SciScore}_F^{IP}$ the fine-tuned model's score under implicit prompts:

\begin{equation}
    RI = \frac{\text{SciScore}_F^{IP}-\text{SciScore}_B^{IP}}{\text{SciScore}_B^{EP}-\text{SciScore}_B^{IP}}
\end{equation}
An RI of 100\% would indicate that fine-tuning has fully closed the gap between implicit and explicit prompts.

\subsection{Two-Stage Training Yields Over 50\% Improvement}
\begin{figure*}[t]
    \centering
    \includegraphics[width=\linewidth]{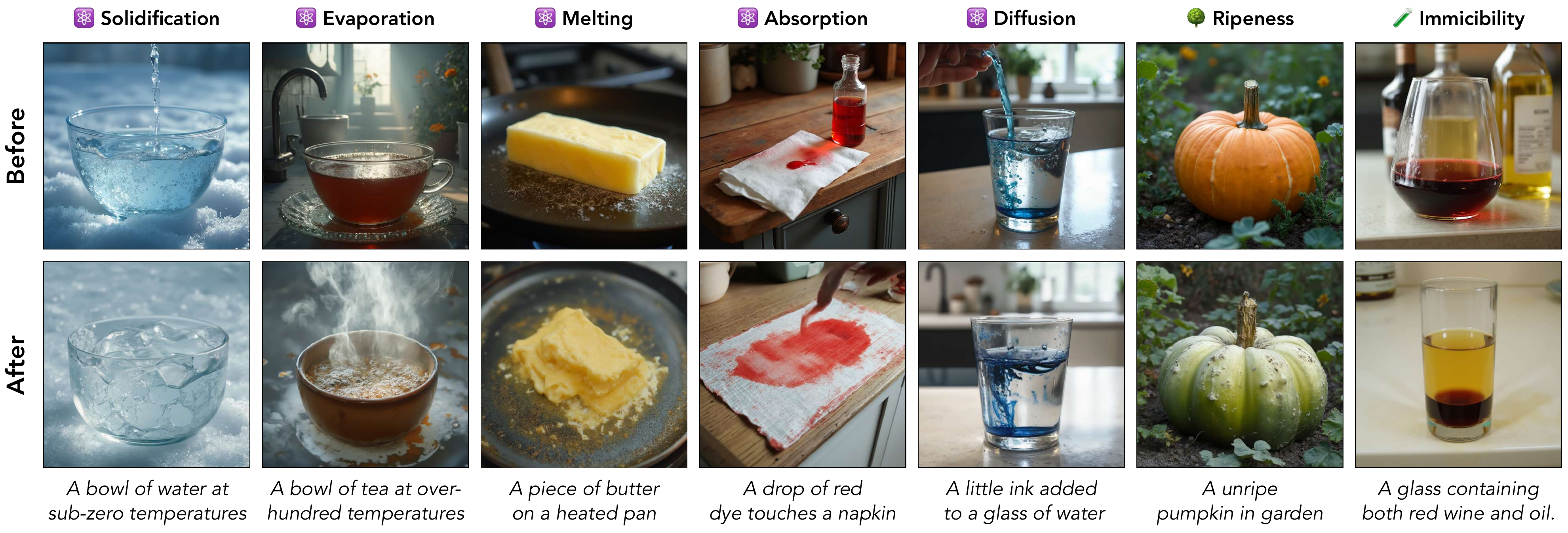}
    \caption{\textbf{The two-stage framework corrects scientific errors while preserving visual quality.} Upper row: base FLUX.1[dev]~\cite{FLUX}. Lower row: our fine-tuned model. Each pair uses an identical random seed. Displayed prompts are simplified summaries.}
    \label{fig:case_study}
    \vspace{-12pt}
\end{figure*}

\paragraph{SFT and OFT together exceed the baseline by over 50\%.}
Table~\ref{tab:results} confirms that both SFT and OFT improve the performance of FLUX~\cite{FLUX} on~\model. The combined two-stage framework increases the score from 23.56 to 28.52 on~\dataset-S (RI = 53.39\%) and from 27.26 to 30.11 on~\dataset-C (RI = 38.31\%). We find that SFT contributes the larger share of the improvement, which is expected: SFT provides the foundational knowledge of what scientific images look like, while OFT refines the model's ability to activate that knowledge from implicit prompts. A qualitative comparison is presented in Figure~\ref{fig:case_study}.

\paragraph{The fine-tuned model generalizes to complex scenes.}
While the~\dataset~training set primarily contains images in straightforward scenarios, the fine-tuned model shows clear improvement on~\dataset-C, which embeds scientific tasks within diverse environmental contexts. This indicates that the model has internalized underlying scientific principles rather than memorizing training examples: the knowledge transfers to visual settings not seen during fine-tuning.

\subsection{Both Stages and Masking Are Necessary}

\begin{table}[t]
  \centering
  \begin{minipage}{0.5\linewidth}
    \centering
    \includegraphics[width=\linewidth]{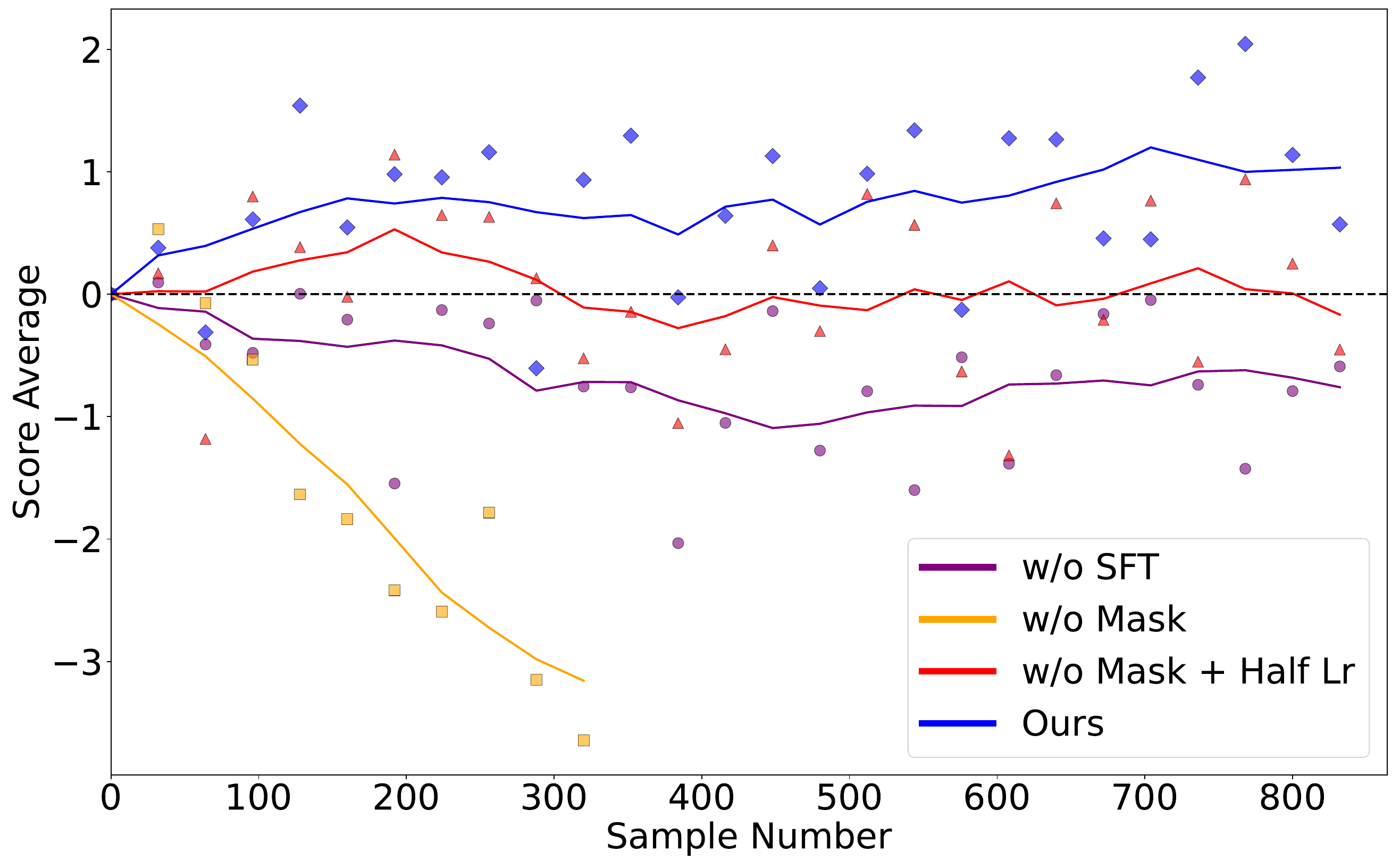}
    \captionof{figure}{\textbf{SFT is essential before OFT, and masking stabilizes training.} Without prior SFT (purple), OFT fails to improve~\model. Without masking (yellow/red), performance becomes erratic or stalls.}
    \label{fig:abla_training}
  \end{minipage}
  \hfill
  \begin{minipage}{0.47\linewidth}
    \centering
     \caption{\textbf{The two-stage framework improves FLUX.1[dev]~\cite{FLUX} by over 50\% in RI.} \textbf{Bold} indicates best performance.}
    \resizebox{\linewidth}{!}{
    \begin{tabular}{l|cc|cc}
        \toprule
        \multirow{2}{*}{Method} & \multicolumn{2}{c|}{\dataset-S} & \multicolumn{2}{c}{\dataset-C} \\
        \cmidrule(lr){2-3} \cmidrule(lr){4-5}
        & \model & RI & \model & RI \\
        \midrule
        FLUX.1[dev]~\cite{FLUX} & $23.56$ & / & $27.26$ & / \\
        $+$EP & $32.85$ & / & $34.70$ & / \\
        $+$SFT & $27.43$ & $41.66$ & $29.49$ & $29.97$\\
        $+$SFT$+$OFT & $28.52$ & \textbf{53.39} & $30.11$ & \textbf{38.31} \\
        \bottomrule
    \end{tabular}
    }
    \label{tab:results}
  \end{minipage}
  \vspace{-12pt}
\end{table}

We ablate each component to verify that both stages and the masking strategy are necessary. Throughout, we follow a consistent protocol: at each training step, all implicit prompts from~\dataset-S are used to generate two images per prompt, and the average~\model~is computed. All curves in Figure~\ref{fig:abla_training} show deviations from the baseline.

\paragraph{OFT without preceding SFT fails to improve performance.}
The blue curve in Figure~\ref{fig:abla_training} shows OFT applied after SFT, while the purple curve shows OFT without prior SFT. Both use identical OFT configurations. We find that SFT followed by OFT yields a stable increase in~\model, whereas OFT alone produces no improvement. Without the scientific knowledge base provided by SFT, the model receives two images that are both scientifically incorrect; the preference signal between two poor samples provides insufficient gradient information for meaningful learning.

\paragraph{Masking suppresses noise from irrelevant visual features.}
Starting from the SFT checkpoint, we test two configurations without masking: standard learning rate (yellow curve) and halved learning rate (red curve). At the standard rate, the model collapses because it tries to match all visual features of the preferred image, including irrelevant background elements. Halving the learning rate prevents collapse but fails to increase~\model, as the noisy gradients from irrelevant regions cancel out the useful signal from the scientific region. In contrast, the masked configuration (blue curve) produces a stable and consistent improvement, confirming that restricting gradients to the scientifically relevant region is essential for effective OFT.

\vspace{-9pt}
\section{Related Work}

\subsection{Scientific Reasoning in Image Generation}

A growing body of work investigates whether generative models can adhere to real-world physical laws, spanning image synthesis~\cite{meng2024phybench,lin2024aglldiff}, video generation~\cite{bansal2024videophy, meng2024phygenbench, kang2024farvideogenerationworld,chai2023stablevideo}, and 3D modeling~\cite{guo2024physically}. In the image domain, PhyBench~\cite{meng2024phybench} introduces a benchmark of physical commonsense tasks and evaluates image generation models through LMM-based scoring. Commonsense-T2I~\cite{fu2024commonsenset2ichallengetexttoimagegeneration} provides paired prompts that test everyday physical reasoning. In the video domain, VideoPhy~\cite{bansal2024videophy} and PhyGenBench~\cite{meng2024phygenbench} assess whether generated videos obey physical dynamics across material interactions. While these works collectively establish that current models lack scientific reasoning, they share two limitations. First, they are evaluation-only: none provides training data that could be used to improve model capabilities. Second, their evaluation protocols rely on LMM-based discrete scoring, which we find unreliable for fine-grained scientific judgment. Our work addresses both gaps by providing a large-scale adversarial training set alongside the test set, and by developing~\model, a specialized reward model that offers continuous, calibrated scoring for scientific correctness.

\subsection{Post-Training Alignment for Diffusion Models}

Post-training methods for diffusion models have been explored along several directions. Rejection-sampling approaches such as VersaT2I~\cite{guo2024versat2i} and DreamSync~\cite{sun2023dreamsync} filter generated outputs to retain higher-quality samples. ReNO~\cite{eyring2024reno} optimizes the initial latent noise at inference time using a differentiable objective. More recently, policy-gradient and preference-based methods have been applied directly to the denoising process~\cite{wallace2024diffusion,yang2024using, black2024trainingdiffusionmodelsreinforcement}, using reward signals derived from DPO~\cite{rafailov2024direct} or PPO~\cite{schulman2017proximalpolicyoptimizationalgorithms}. These methods share a common assumption: the desired outputs already lie within the pre-trained model's generative distribution, so optimization amounts to steering the model toward preferred regions of that distribution. This assumption holds for aesthetic or compositional preferences, where the model has already seen relevant visual patterns during pretraining. However, it breaks down for scientific reasoning, where the target phenomena may be entirely absent from the pretraining data. Our framework addresses this by introducing an SFT stage that first exposes the model to scientific visual patterns, followed by online fine-tuning with a domain-specific reward model to optimize for implicit reasoning.

\subsection{Benchmarking Image Generation}

Standard metrics such as FID~\cite{heusel2017gans}, IS~\cite{salimans2016improved}, LPIPS~\cite{zhang2018unreasonable}, and CLIPScore~\cite{hessel2021clipscore} remain widely used but primarily measure distributional similarity or general text-image alignment. Recent work has introduced richer evaluation paradigms: HPSv2~\cite{wu2023human}, PickScore~\cite{kirstain2023pickapicopendatasetuser}, and ImageReward~\cite{xu2023imagereward} capture human aesthetic preferences; VQA-based methods such as VQAScore~\cite{li2024evaluating}, TIFA~\cite{hu2023tifa}, VIEScore~\cite{ku2023viescore}, LLMscore~\cite{lu2023llmscore}, and DSG~\cite{cho2023davidsonian} probe compositional fidelity through question answering; and T2I-CompBench~\cite{huang2023t2i} and CLIP-R-Precision~\cite{park2021benchmark} target object attributes and spatial relationships.

However, most of these benchmarks focus on visual aesthetics and compositional fidelity rather than scientific correctness. PhyBench~\cite{meng2024phybench} and Commonsense-T2I~\cite{fu2024commonsenset2ichallengetexttoimagegeneration} take important steps toward evaluating physical and commonsense reasoning, but they provide only test prompts and rely on discrete LMM-based scoring. We introduce~\model, a reward model that provides continuous, fine-grained scoring specifically calibrated for scientific correctness, and pair it with a large-scale training set to enable both evaluation and model improvement.

\section{Conclusion}

In this work, we investigate the gap between visual fidelity and scientific realism in image generation. We construct~\dataset, an expert-annotated dataset of over 20k adversarial image pairs spanning 16 scientific domains, and use its test set to benchmark 18 recent models. We find that no model scores above 50 out of 100 under implicit scientific prompts, while the same models score roughly 35 points higher when given explicit descriptions of the intended outcome. This confirms that the bottleneck is not in visual rendering but in scientific reasoning.

To address this gap, we develop~\model, a reward model fine-tuned from CLIP-H~\cite{ilharco_gabriel_2021_5143773} that surpasses state-of-the-art LMMs and experienced human evaluators by roughly 5 points on our benchmark. Building on~\model, we propose a two-stage alignment framework: supervised fine-tuning first provides the scientific knowledge base, and masked online fine-tuning then refines the model's implicit reasoning ability using~\model~as the reward signal. Applying this framework to FLUX.1[dev]~\cite{FLUX} yields over 50\% relative improvement, with the fine-tuned model generalizing to complex scenes not seen during training.

Collectively, these results demonstrate that scientific reasoning in image generation is not an inherent limitation of current architectures but a consequence of missing data and misaligned training objectives. We hope that~\dataset~and~\model~will serve as useful resources for future research on grounding generative models in reality.

\section*{Acknowledgments}
We thank Alistair King for sharing insightful code, which was instrumental in our fine-tuning process. SX also acknowledges support from Open Path AI Foundation, Intel AI SRS, IITP grant funded by the Korean Government (MSIT) (No. RS-2024-00457882, National AI Research Lab Project), Amazon Research Award, and NSF Award IIS-2443404.

\bibliographystyle{unsrtnat}
\bibliography{references}

\newpage
\appendix
\onecolumn

\makeappendixtitle

\noindent The Appendix is structured as follows:
\begin{itemize}[leftmargin=7.5mm]
\setlength{\itemsep}{3pt}
\item Section~\ref{sec:supp_rewrite_cap}: why we restrict tasks to linguistically describable phenomena, and why tasks such as reflections and shadows are excluded.
\item Section~\ref{sec:supp_tasks}: definitions and visual contrasts for all 16 tasks, organized by Biology, Chemistry, and Physics.
\item Section~\ref{sec:supp_obs}: the subject + condition decomposition that underlies ST/CT classification, with the full task assignment.
\item Section~\ref{sec:supp_dataset_detail}: three-stage data curation pipeline covering prompt generation, image synthesis, and quality control.
\item Section~\ref{sec:supp_bench_comp}: a comparison of~\dataset~with existing scientific and commonsense reasoning benchmarks.
\item Section~\ref{sec:supp_train_sciscore}: hyper-parameters for training~\model.
\item Section~\ref{sec:supp_baseline_sciscore}: evaluation protocol and setup for VLM, LMM, and human baselines.
\item Section~\ref{sec:supp_result_sciscore}: per-category accuracy breakdowns on both test splits, revealing where~\model~achieves perfect accuracy and where it struggles.
\item Section~\ref{sec:supp_analysis_sciscore}: analysis of why VLMs approach random guessing, why LMMs fail despite CoT, and why~\model~surpasses human evaluators.
\item Section~\ref{sec:supp_abla_sciscore}: qualitative examples showing how IEE improves fine-grained visual discrimination.
\item Section~\ref{sec:supp_two_stage_train_detail}: mathematical derivations for SFT, DPO, noise scheduling, subject extraction, and gradient masking.
\item Section~\ref{sec:supp_ft_train_setting}: hyper-parameter configurations for both the SFT and OFT stages.
\item Section~\ref{sec:supp_res_oft}: additional OFT experiments using LAION aesthetic and ImageReward as alternative reward models.
\item Section~\ref{sec:supp_limitation}: discussion of cross-task generalizability limitations and subject-oriented task challenges.
\end{itemize}

\section{Why We Prioritize Linguistically Describable Tasks}
\label{sec:supp_rewrite_cap}

A natural question when designing~\dataset~is why we restrict the task set to phenomena whose visual outcomes can be fully described through text. During the design process, we considered several additional reasoning categories that would broaden scientific coverage. For instance, reflection-based tasks~\cite{farid2022perspectiveinconsistencypainttext} evaluate whether objects and their mirror reflections are geometrically consistent, and shadow-based tasks~\cite{sarkar2024shadowsdontlielines} test whether cast shadows follow the correct light-source geometry.

However, we find that these tasks present a fundamental obstacle for our adversarial data curation pipeline. Both correct and incorrect versions of a reflection or shadow require precise control over geometric details that cannot be specified through text alone. Current image generation models lack the spatial precision to reliably produce such controlled pairs, making it impossible to construct the tuples that~\dataset~relies on for preference modeling.

This observation leads to a deliberate design principle: every task in~\dataset~must satisfy the \textit{rewriting capability} requirement, meaning that the scientific visual outcome can be unambiguously expressed through an alternative textual description. For example, ``an unripe apple'' can be rewritten as ``a green apple,'' preserving the intended visual meaning. This property guarantees that (1) explicit and superficial prompt variants can be derived from any implicit prompt, and (2) image generation models can interpret and produce the target images reliably. Tasks involving geometric or lighting precision beyond the reach of textual prompts, such as reflections~\cite{farid2022perspectiveinconsistencypainttext} and subtle illumination effects~\cite{IC-Light}, remain important directions for future work as generation models improve in spatial control. 
\section{Task Definitions and Visual Contrasts}
\label{sec:supp_tasks}

Each of the 16 tasks in~\dataset~targets a specific scientific phenomenon and defines the visual contrast that the model must reason about. We organize them by scientific domain below. Illustrative examples from the dataset are shown in Figures~\ref{fig:case1},~\ref{fig:case2}, and~\ref{fig:case3}.

\begin{itemize}
    \item \textbf{Light Requirement~(LR)}: Plants change color and leaf size depending on whether they receive adequate or insufficient light.
    \item \textbf{Watering Requirement~(WR)}: Plants exhibit differences in foliage health and growth under sufficient versus inadequate watering, with the latter causing wilting and reduced leaf size.
    \item \textbf{Ripeness~(RI)}: Fruits alter their color and texture when ripe compared to unripe.
    \item \textbf{Seasonal Change~(SC)}: Plants display variations in leaf color, size, and blooming patterns across seasons.
    \item \textbf{Flame Reaction~(FR)}: Chemical substances produce characteristic flame colors when burned.
    \item \textbf{Immiscibility~(IM)}: Two liquids either mix uniformly or separate into visible layers based on chemical properties.
    \item \textbf{Rust~(RU)}: Metals appear shiny and reflective before oxidation, and corroded, flaky, and discolored after rusting.
    \item \textbf{Absorption~(AB)}: A solid either absorbs a liquid or repels it, depending on their material properties.
    \item \textbf{Buoyancy~(BU)}: Objects either float on or sink in water based on their density relative to water.
    \item \textbf{Diffusion~(DI)}: A small amount of liquid added to another either disperses uniformly or remains separate.
    \item \textbf{Electricity~(EL)}: Electronic devices change appearance (e.g., glowing, sparking) when electric current is applied versus disconnected.
    \item \textbf{Evaporation~(EV)}: Liquids boil and produce vapor upon reaching boiling point; below it, the surface remains calm.
    \item \textbf{Gravity~(GR)}: Objects appear differently positioned under normal gravity versus a gravity-free environment.
    \item \textbf{Liquidation~(LI)}: Air condenses into water droplets on surfaces cooled below room temperature.
    \item \textbf{Melting~(ME)}: Solids transition to liquid, changing shape and structure upon reaching their melting point.
    \item \textbf{Solidification~(SO)}: Liquids become solids, altering form and texture when cooled below their solidification point.
\end{itemize}

\begin{figure}[t]
    \centering
    \includegraphics[width=0.98\linewidth]{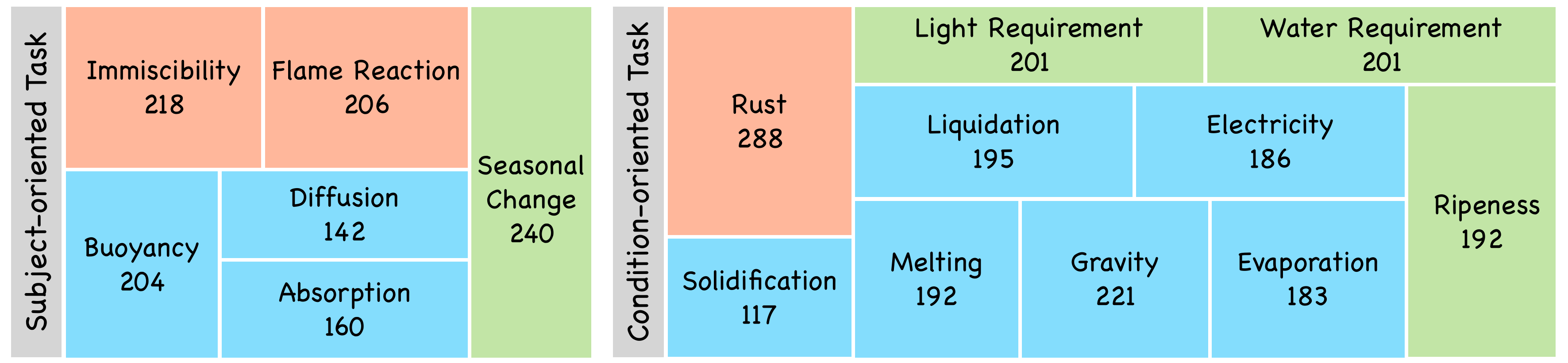}
    \caption{\textbf{Subject-oriented vs.\ condition-oriented task classification.} Ten of the 16 tasks in~\dataset~are condition-oriented and six are subject-oriented. Condition-oriented tasks involve generalizable visual patterns, while subject-oriented tasks require subject-specific scientific knowledge.}
    \label{fig:task_cls}
    \vspace{-6pt}
\end{figure}

\section{All Tasks Follow a Subject + Condition Structure}
\label{sec:supp_obs}

We observe that every task in~\dataset~can be decomposed into a \textbf{subject} and a \textbf{condition}. For example, the prompt \textit{an unripe apple} comprises the subject \textit{apple} and the condition \textit{unripe}; \textit{a laptop without electricity} includes the subject \textit{laptop} and the condition \textit{without electricity}. This uniform structure motivates a two-way classification based on \textit{where} the scientific reasoning lies: in the subject or in the condition.

\paragraph{Subject-oriented tasks.}
In these tasks, the need for scientific reasoning arises from the subject's intrinsic properties. Different subjects under the same condition exhibit different visual features. For example, the \textit{buoyancy} task is subject-oriented because whether an object floats or sinks depends on its density relative to water, an intrinsic property that varies across subjects. Other subject-oriented tasks include \textit{flame reaction} (different metals produce different flame colors), \textit{immiscibility}, \textit{absorption}, \textit{diffusion}, and \textit{electricity}.

\paragraph{Condition-oriented tasks.}
In these tasks, the scientific reasoning is tied to the applied condition. Varying subjects under a single condition produce similar visual outcomes. For instance, the \textit{gravity} task is condition-oriented because most subjects behave similarly: floating in zero gravity and resting on the ground under normal gravity. The remaining condition-oriented tasks include \textit{ripeness}, \textit{seasonal change}, \textit{light requirement}, \textit{watering requirement}, \textit{rust}, \textit{evaporation}, \textit{liquidation}, \textit{melting}, and \textit{solidification}.

\paragraph{Why this distinction matters.}
As shown in Figure~\ref{fig:task_cls}, ten tasks are condition-oriented and six are subject-oriented. This classification proves analytically useful: as we demonstrate in Section~\ref{sec:rw_result}, nearly all failure cases of~\model~concentrate in subject-oriented tasks, where the model must have prior exposure to each subject's specific properties. Condition-oriented tasks, by contrast, involve generalizable visual patterns that transfer more easily across subjects.

\vspace{-6pt}
\section{Data Curation Pipeline: Prompts, Images, and Quality Control}
\label{sec:supp_dataset_detail}

In this section, we describe the three stages of our data curation pipeline in detail.

\paragraph{Stage 1: prompt generation.}
For each of the 16 tasks, we use \texttt{GPT-4o}~\cite{GPT} to define a set of structured templates for implicit prompts. These templates capture the core scientific principle while allowing variability in the objects or substances involved. Using these templates, \texttt{GPT-4o} generates a diverse set of implicit prompts by inserting appropriate subjects. For each implicit prompt, \texttt{GPT-4o} then produces the corresponding explicit and superficial prompts following the instruction template shown in Figure~\ref{fig:prompt_col}. This three-stage process (template $\rightarrow$ implicit $\rightarrow$ explicit/superficial) ensures that all three prompt types within a tuple remain semantically aligned while differing precisely in their level of scientific specificity.

\begin{figure}[t]
\centering
\begin{tcolorbox}[title=User Prompt]
\small Assume you are an experienced scientist. Your task is to generate both an explicit prompt and a superficial prompt based on a given input prompt. The input prompt is formulated with scientific principles and will serve as input for a text-to-image generative model. It may include terminology or phrases that are not overtly descriptive but imply certain visual characteristics or phenomena, requiring interpretative scientific reasoning to convey their meaning.\\[6pt]
\textbf{Explicit Prompt:} Reformulate the input prompt into a precise, descriptively accurate statement that aligns with the intended visual outcome, incorporating the implied scientific nuances.\\[6pt]
\textbf{Superficial Prompt:} Construct an interpretation that disregards the underlying scientific reasoning. Focus only on the superficial or literal descriptive aspects.\\[6pt]
\textbf{Example:} \{``input prompt": ``an unripe apple", ``explicit prompt": ``a green apple", ``superficial prompt": ``a red apple"\}\\[6pt]
Here is the input prompt: [Your Input]. Please output in the following format:\\[3pt]
\{``explicit prompt": , ``superficial prompt": \}
\end{tcolorbox}
\vspace{-9pt}
\caption{\textbf{Instruction template for prompt generation.} Given an implicit prompt, \texttt{GPT-4o}~\cite{GPT} generates the corresponding explicit and superficial prompts.}
\label{fig:prompt_col}
\vspace{-9pt}
\end{figure}

\begin{figure*}[h]
    \centering
    \includegraphics[width=0.98\linewidth]{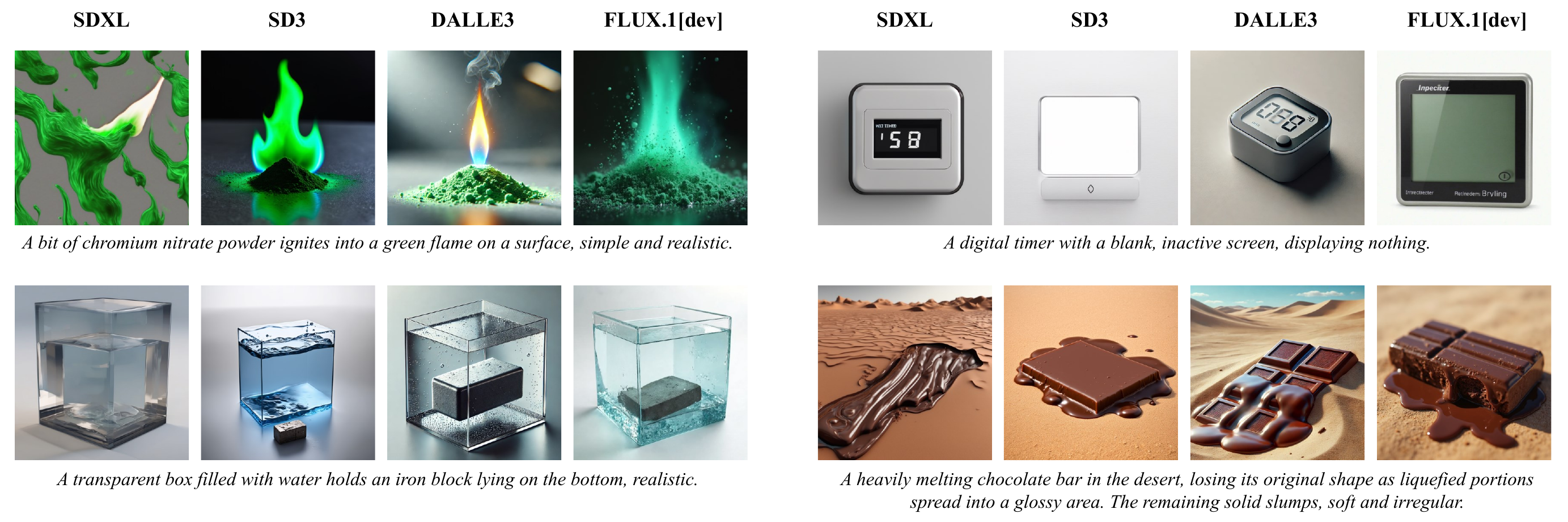}
    \caption{\textbf{FLUX.1[dev] produces the most faithful and realistic images among compared models.} SDXL~\cite{podell2023sdxlimprovinglatentdiffusion}, SD~3~\cite{sd3}, and DALLE~3~\cite{dalle} occasionally fail to align generated images with the provided descriptions, while FLUX.1[dev]~\cite{FLUX} consistently renders accurate content with photorealistic quality.}
    \label{fig:model_choice}
\end{figure*}

\paragraph{Stage 2: image synthesis.}
Images relevant to our scientific reasoning tasks are scarce in existing datasets. We therefore generate synthetic images, selecting the generation model based on two criteria:
\begin{itemize}
    \item \textbf{Text-image alignment.} The model must accurately render both explicit and superficial prompts, as faithful alignment is essential for constructing valid adversarial pairs. Misalignment at this stage would introduce label noise into the training set.
    \item \textbf{Photorealistic style.} Because our tasks are grounded in real-world phenomena, the generated images must exhibit a photorealistic style. Abstract or stylized renderings would obscure the fine-grained visual differences that distinguish scientifically correct from incorrect images.
\end{itemize}
We compare SDXL~\cite{podell2023sdxlimprovinglatentdiffusion}, SD~3~\cite{sd3}, DALLE~3~\cite{dalle}, and FLUX.1[dev]~\cite{FLUX} on a representative subset of prompts. As shown in Figure~\ref{fig:model_choice}, FLUX.1[dev] consistently produces the most faithful and realistic outputs across all task categories. We therefore adopt FLUX.1[dev] for all image generation in~\dataset.

\begin{figure*}[t]
    \centering
    \subfloat{
        \includegraphics[width=0.98\linewidth]{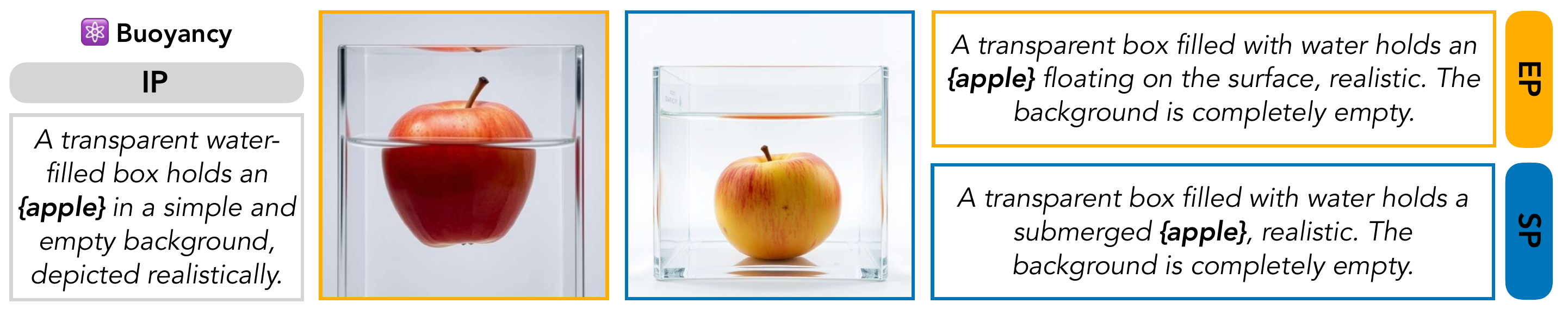}
    }
    \hfill
    \vspace{6pt}
    \subfloat{
        \includegraphics[width=0.98\linewidth]{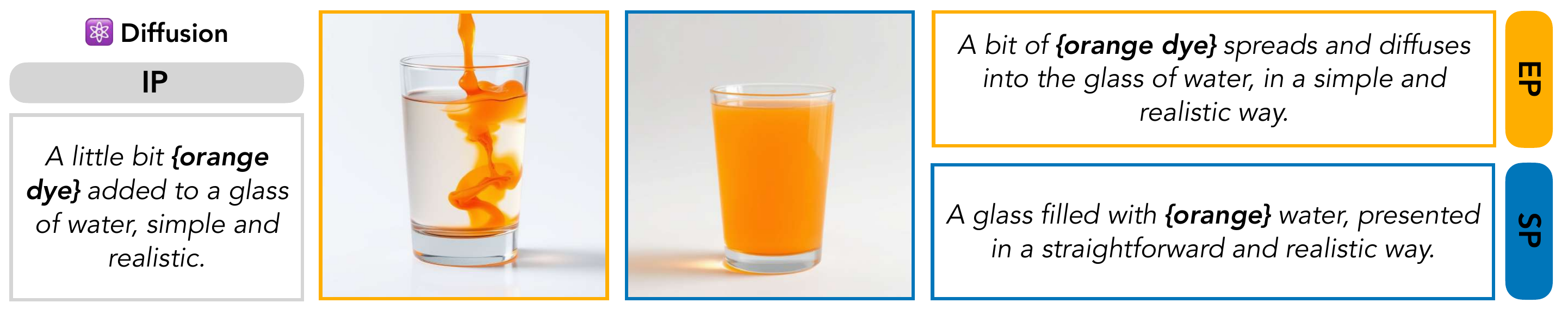}
    }
    \hfill
    \vspace{6pt}
    \subfloat{
        \includegraphics[width=0.98\linewidth]{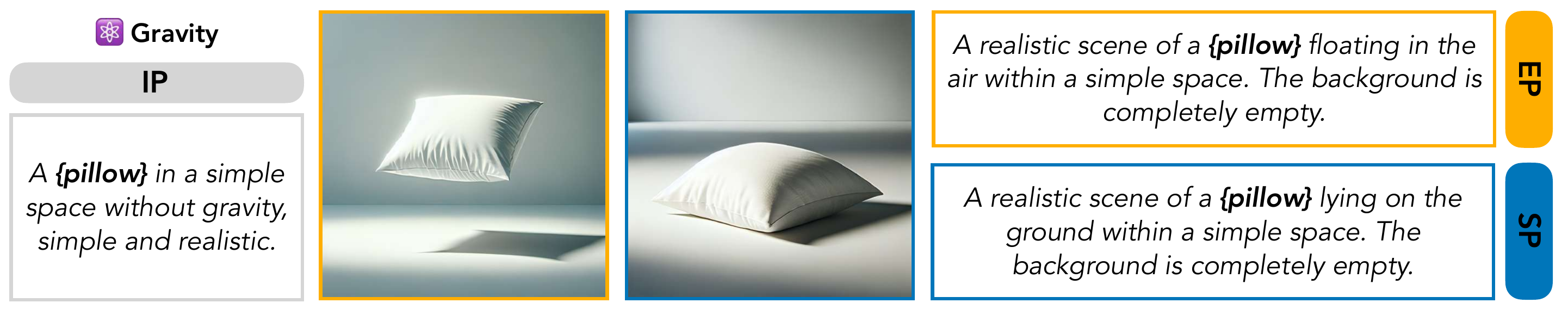}
    }
    \hfill
    \vspace{6pt}
    \subfloat{
        \includegraphics[width=0.98\linewidth]{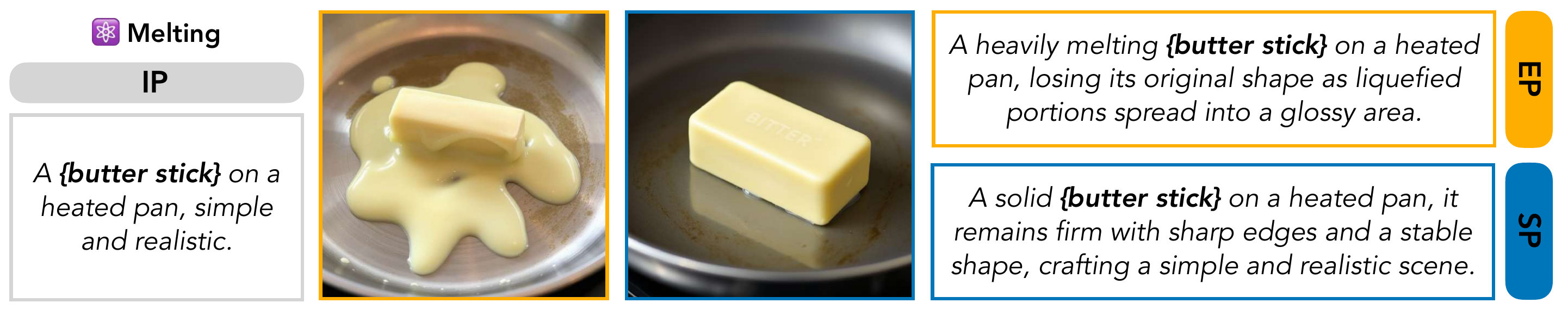}
    }
    \hfill
    \vspace{6pt}
    \subfloat{
        \includegraphics[width=0.98\linewidth]{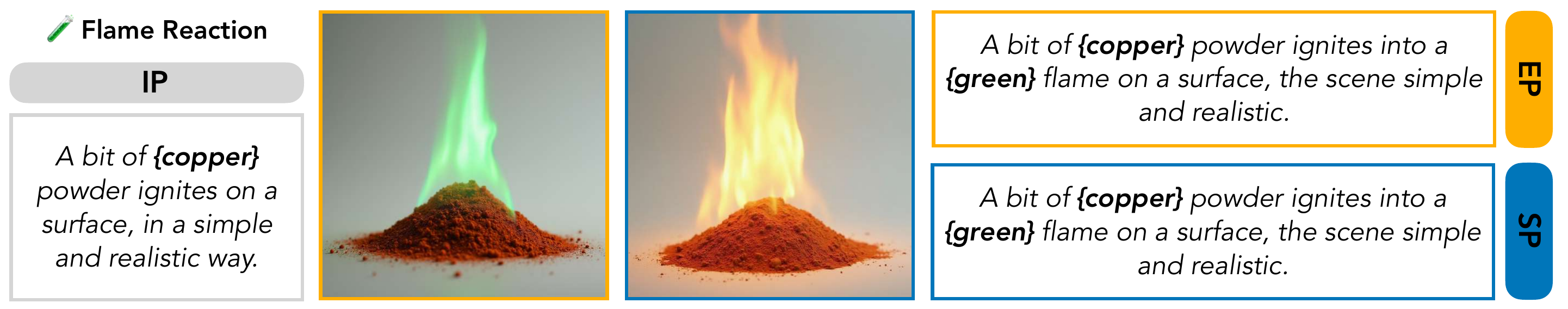}
    }
    \hfill
    \vspace{6pt}
    \subfloat{
        \includegraphics[width=0.98\linewidth]{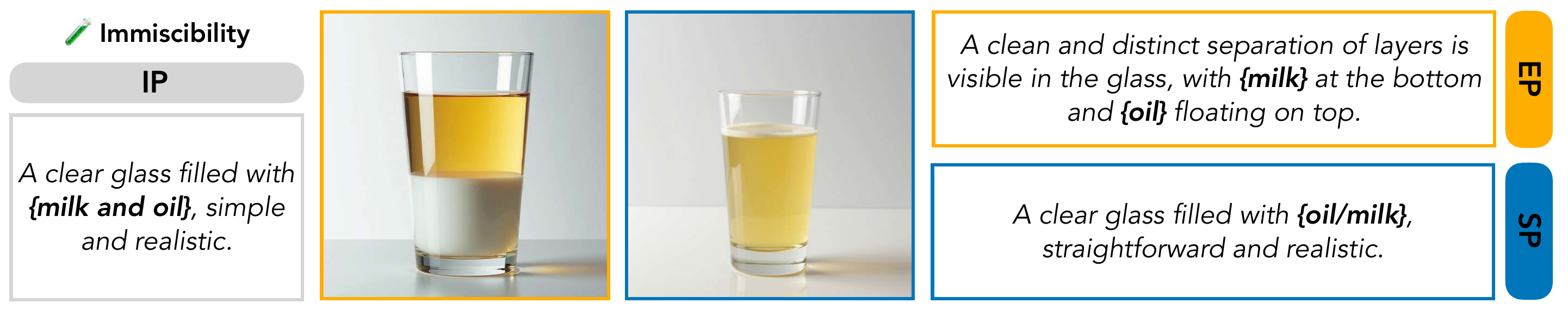}
    }
    \caption{\textbf{Several examples from~\dataset.} 'EP' denotes explicit prompts (yellow blocks), 'SP' denotes superficial prompts (blue blocks), and 'IP' denotes implicit prompts (grey blocks).}
    \label{fig:case1}
\end{figure*}

\begin{figure*}[t]
    \centering
    \subfloat{
        \includegraphics[width=0.98\linewidth]{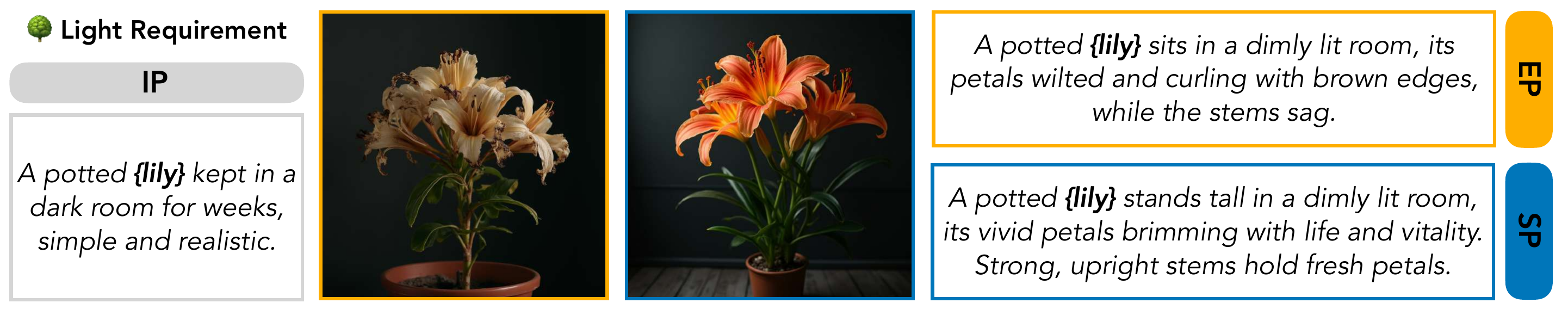}
    }
    \hfill
    \vspace{6pt}
    \subfloat{
        \includegraphics[width=0.98\linewidth]{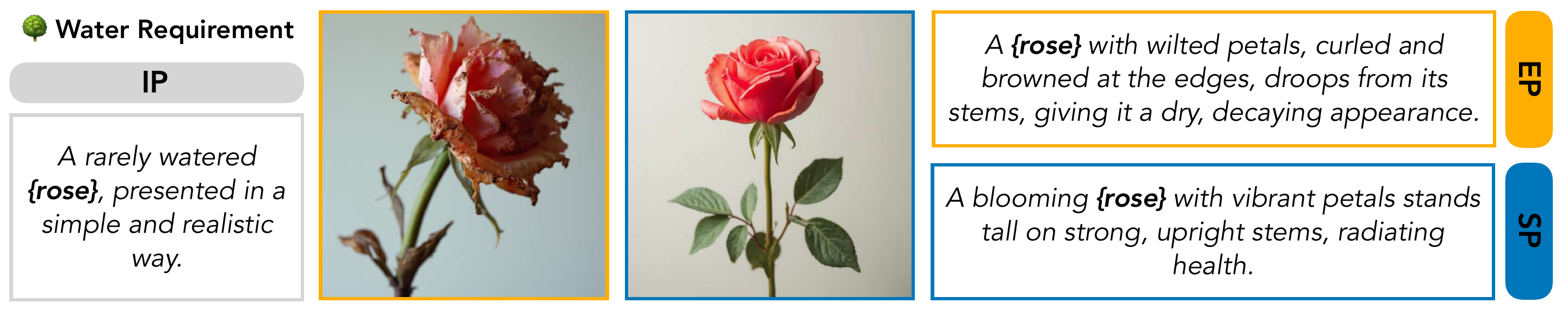}
    }
    \hfill
    \vspace{6pt}
    \subfloat{
        \includegraphics[width=0.98\linewidth]{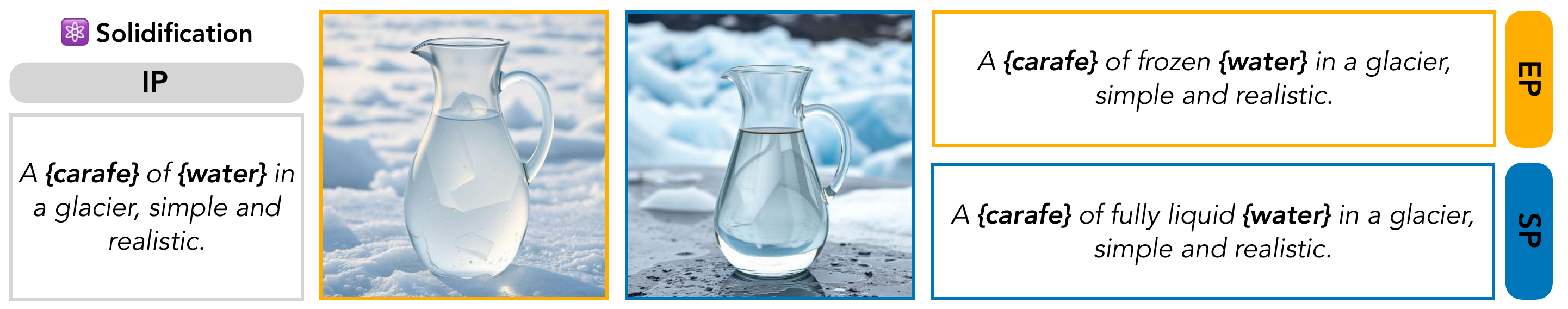}
    }
    \hfill
    \vspace{6pt}
    \subfloat{
        \includegraphics[width=0.98\linewidth]{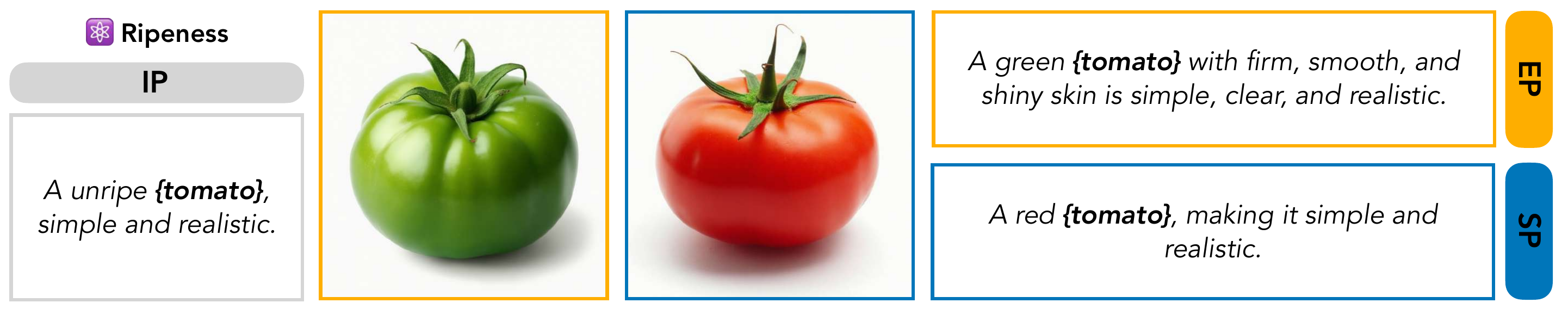}
    }
    \hfill
    \vspace{6pt}
    \subfloat{
        \includegraphics[width=0.98\linewidth]{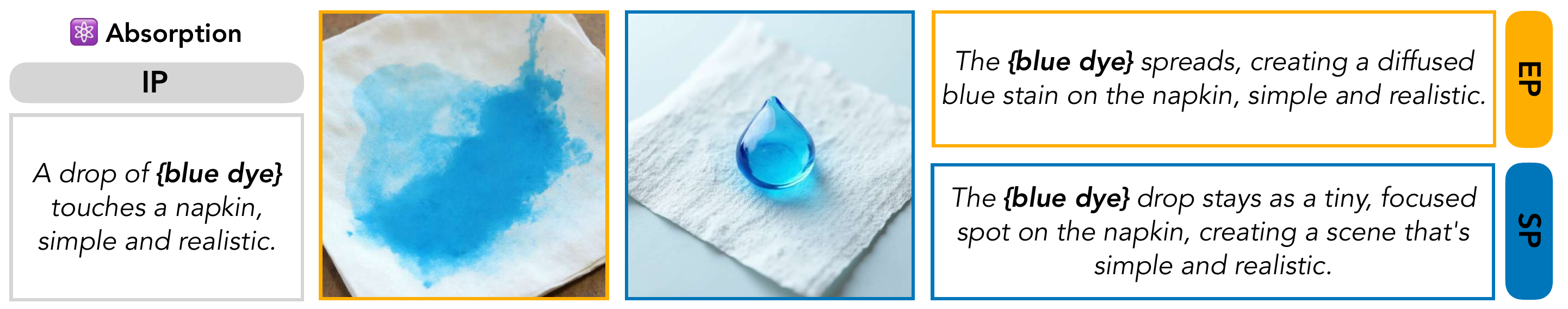}
    }
    \hfill
    \vspace{6pt}
    \subfloat{
        \includegraphics[width=0.98\linewidth]{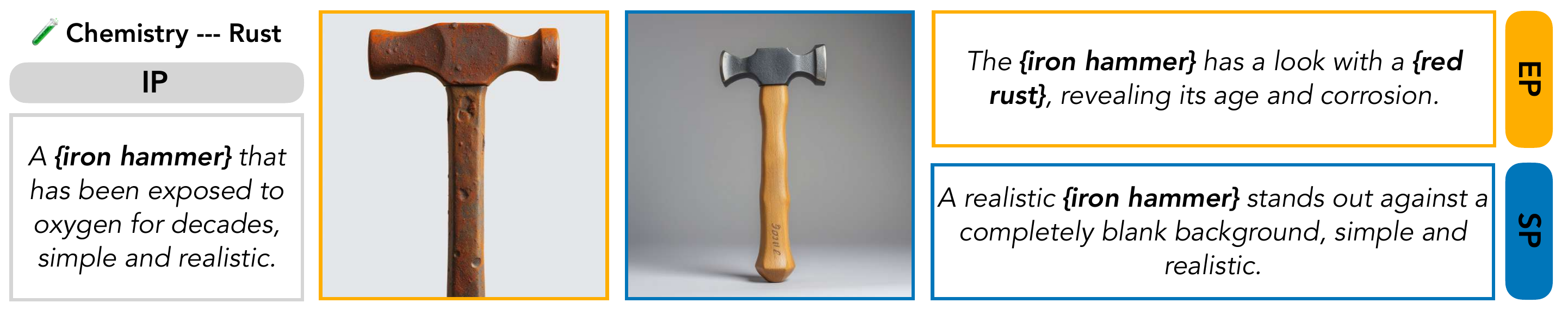}
    }
     \caption{\textbf{Several examples from~\dataset.} 'EP' denotes explicit prompts (yellow blocks), 'SP' denotes superficial prompts (blue blocks), and 'IP' denotes implicit prompts (grey blocks).}
    \label{fig:case2}
\end{figure*}

\begin{figure*}[t]
    \centering
    \subfloat{
        \includegraphics[width=0.98\linewidth]{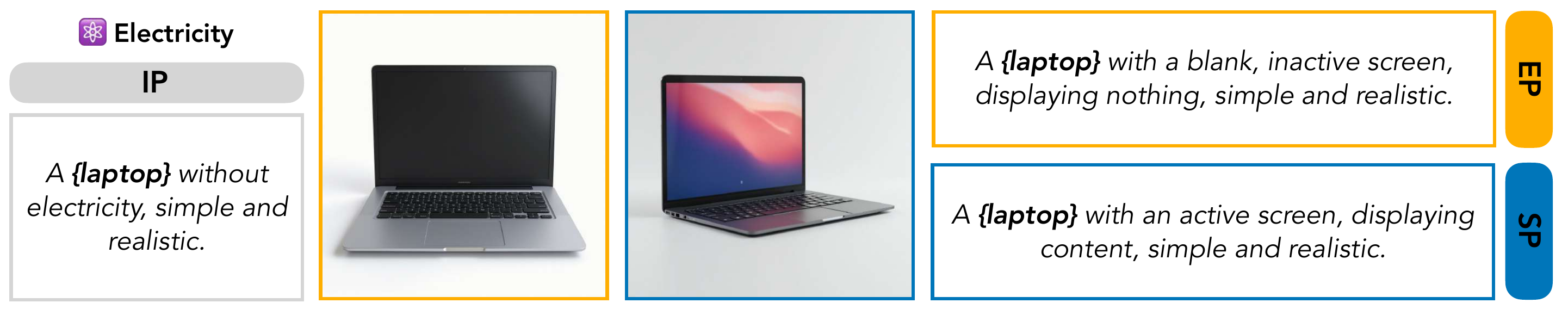}
    }
    \hfill
    \vspace{6pt}
    \subfloat{
        \includegraphics[width=0.98\linewidth]{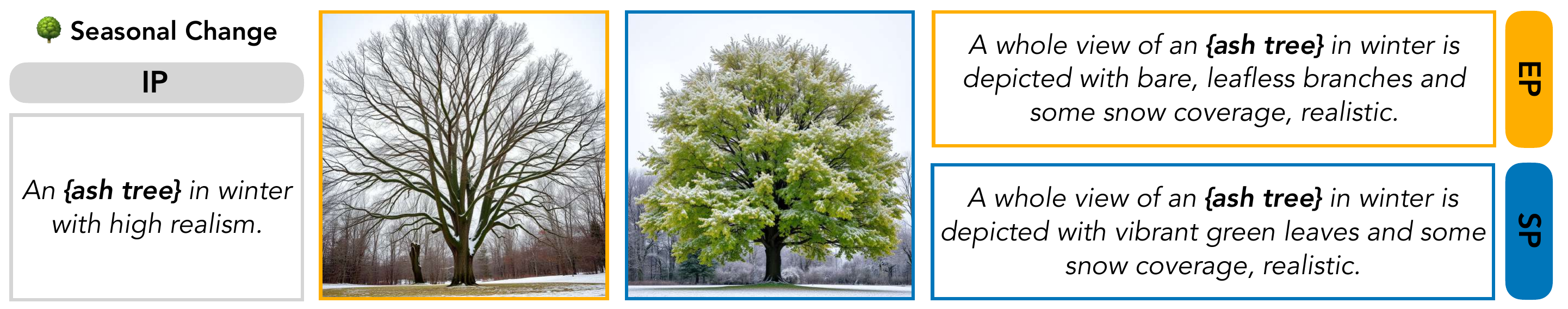}
    }
    \hfill
    \vspace{6pt}
    \subfloat{
        \includegraphics[width=0.98\linewidth]{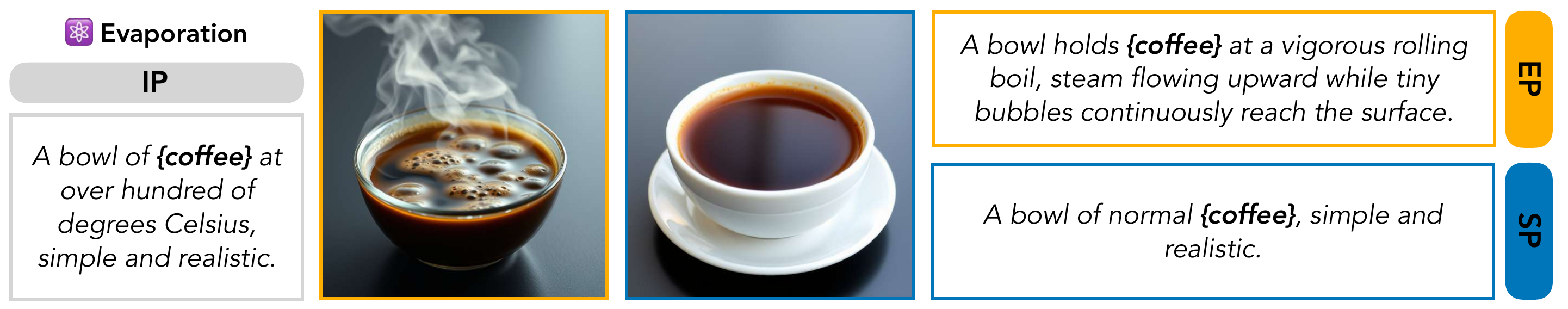}
    }
    \hfill
    \vspace{6pt}
    \subfloat{
        \includegraphics[width=0.98\linewidth]{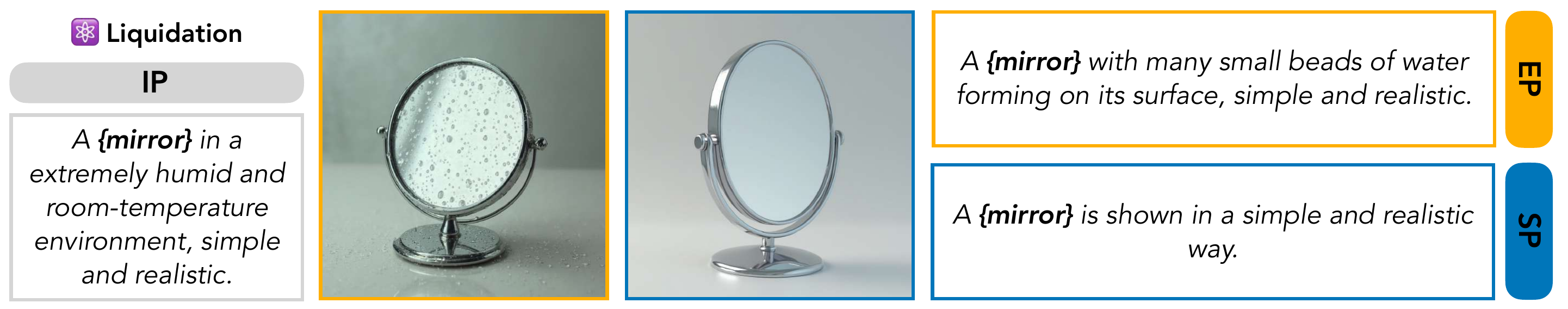}
    }
    \caption{\textbf{Several examples from~\dataset.} 'EP' denotes explicit prompts (yellow blocks), 'SP' denotes superficial prompts (blue blocks), and 'IP' denotes implicit prompts (grey blocks).}
    \label{fig:case3}
\end{figure*}

\paragraph{Stage 3: quality control.}
The scientific principles embedded in each implicit prompt manifest as specific visual features in localized regions of the image (e.g., the color of a flame, the surface texture of a rusted metal). During curation, we focus on ensuring these regions accurately reflect the underlying science. We apply two filtering criteria:
\begin{itemize}
    \item \textbf{Background simplicity.} We select images with minimal background complexity (e.g., solid colors) to eliminate distracting visual elements that could confound the scientific signal.
    \item \textbf{Region prominence.} We filter to ensure that the scientifically relevant region occupies a large portion of the image, maximizing the visual salience of the target phenomenon.
\end{itemize}
Representative examples of curated images are shown in Figures~\ref{fig:case1},~\ref{fig:case2}, and~\ref{fig:case3}. All generated tuples undergo a final round of manual verification by domain experts, who cross-reference each image against established scientific knowledge to confirm correctness.

\section{\dataset~Unifies Training and Dual-Purpose Evaluation}
\label{sec:supp_bench_comp}

Table~\ref{tab:benchmark} compares~\dataset~with existing benchmarks targeting physical or commonsense reasoning in image generation. Most prior benchmarks provide only test prompts and support only one evaluation mode. In contrast,~\dataset~is designed around two complementary capabilities.

First,~\dataset~includes a training set of over 20k adversarial pairs, enabling both reward model training and generative model alignment. To our knowledge, none of the existing benchmarks in this space provides a training set.

Second, the test set supports two distinct evaluation modes within a single framework. On one hand, it benchmarks the scientific reasoning of image generation models by scoring their outputs under implicit prompts (Section~\ref{sec:benchmark}). On the other hand, the same test set evaluates the scientific judgment of LMMs and VLMs through a two-choice selection task that tests whether these models can distinguish scientifically correct images from incorrect ones (Section~\ref{sec:rw_result}). This dual-purpose design allows researchers to assess both generation and understanding capabilities using the same prompts and grading criteria, eliminating the need for separate benchmarks.

\begin{table}[h]
  \caption{\textbf{\dataset~is the only benchmark that provides a training set and supports both generation and LMM evaluation.} Existing benchmarks offer test-only evaluation for a single task type, while~\dataset~unifies training data, generation benchmarking, and LMM evaluation in one package.}
  \label{tab:benchmark}
  \centering
  \resizebox{0.8\linewidth}{!}{
  \begin{tabular}{@{}l|c|c|c|cc@{}}
    \toprule
         \multirow{2}{*}{Benchmark} & \multirow{2}{*}{Type} & \multirow{2}{*}{Category} & \multirow{2}{*}{Training Set} & \multicolumn{2}{c}{Evaluation} \\ 
   \cmidrule(lr){5-6}
    & & & & {Generation} & {LMM} \\
    \midrule
    Commonsense-T2I~\cite{fu2024commonsenset2ichallengetexttoimagegeneration} & Commonsense & $5$ & \ding{56} & \ding{52} & \ding{56} \\
    T2I-FactualBench~\cite{huang2024t2ifactualbenchbenchmarkingfactualitytexttoimage} & Commonsense & $8$ & \ding{56} & \ding{52} & \ding{56} \\
    PhyBench~\cite{meng2024phybench} & Science & $31$ & \ding{56} & \ding{52} & \ding{56} \\
    \midrule
    \dataset~(Ours) & Science & $16$ & \ding{52} & \ding{52} & \ding{52}\\
    \bottomrule
  \end{tabular}
  }
\end{table}
\section{Training Settings for~\model}
\label{sec:supp_train_sciscore}

Table~\ref{tab:sciscore_hyperparameters} summarizes the hyper-parameters used to train~\model.

\begin{table}[h]
    \centering
    \vspace{-3pt}
    \caption{\textbf{Hyper-parameters for training~\model.}}
    \vspace{-3pt}
    \begin{tabular}{l|c}
        \toprule
        Hyper-parameters & \model \\
        \midrule
        batch size & $128$  \\
        learning rate & $2 \times 10^{-6}$ \\
        learning rate schedule & cosine  \\
        weight decay & $0.3$ \\
        training steps & $600$ \\
        warmup steps & $150$ \\
        optimizer & AdamW~\cite{loshchilov2019decoupledweightdecayregularization} \\
        $\lambda$ & $0.25$ \\
        \bottomrule
    \end{tabular}
    \label{tab:sciscore_hyperparameters}
\end{table}

\vspace{-12pt}
\section{Baseline Setup and Evaluation Process for~\model}
\label{sec:supp_baseline_sciscore}

\paragraph{Evaluation protocol.}
Each evaluation instance presents one implicit prompt alongside two images: one generated from the explicit prompt and one from the superficial prompt. Models and human evaluators are asked to select the image that better corresponds to the implicit prompt. 

\paragraph{Vision-language models (VLMs).}
We evaluate CLIP-H~\cite{ilharco_gabriel_2021_5143773}, BLIP-2~\cite{li2023blip2bootstrappinglanguageimagepretraining}, and SigLIP~\cite{zhai2023sigmoidlosslanguageimage}. For each model, we encode the implicit prompt and both images using their respective text and image encoders, then apply the scoring mechanism described in Section~\ref{sec:reward_model} to select the higher-scoring image.

\paragraph{Large multimodal models (LMMs).}
We evaluate LLaVA-OV~\cite{li2024llavaonevisioneasyvisualtask}, Qwen2-VL~\cite{wang2024qwen2vlenhancingvisionlanguagemodels}, InternVL~\cite{chen2024internvlscalingvisionfoundation}, and \texttt{GPT-4o-mini}~\cite{GPT}. For \texttt{GPT-4o-mini}, we test both a standard setting and a CoT setting~\cite{wei2023chainofthoughtpromptingelicitsreasoning} that encourages step-by-step reasoning. Each LMM is prompted to choose between two images. To mitigate order bias, we evaluate each pair twice with reversed image order and average the accuracy. The full instruction template is shown in Figure~\ref{fig:gpt_eval}.

\begin{figure}[t]
\centering
\begin{tcolorbox}[title=LMM Evaluation Instruction]
\small You will be presented with a textual prompt followed by two visual images. Your task is to critically analyze and compare both images, selecting the one that most accurately aligns with and represents the overall meaning of the given prompt. \textcolor{red}{First, you should imagine how an ideal image would look based on the prompt, and then describe both images in detail. Finally, combining your initial visualization with the descriptions of the two images, you should select the image that most effectively conveys the intended meaning of the prompt, providing a reasoned justification for your choice.}\\[6pt]
\textbf{Input:} \{``prompt": [Your Input Prompt], ``image-1": [Your Input Image], ``image-2": [Your Input Image]\}\\[6pt]
\textbf{Output format:}\\[3pt]
\{\textcolor{red}{`imagination': , `description of image-1': , `description of image-2': , `justification for choice': , }`final choice': \}
\end{tcolorbox}
\vspace{-8pt}
\caption{\textbf{LMM evaluation instruction.} Red text indicates additional fields used for CoT~\cite{wei2023chainofthoughtpromptingelicitsreasoning} reasoning.}
\label{fig:gpt_eval}
\vspace{-12pt}
\end{figure}

\paragraph{Human evaluation.}
We recruit 10 evaluators, all holding at least a bachelor's degree in science or engineering. Unlike the dataset curators, who verify accuracy using scientific literature, the evaluators rely solely on their own knowledge to select the better image. This distinction explains why human performance, while strong, is not perfect.

\vspace{-12pt}
\section{Per-Category Results of~\model}
\label{sec:supp_result_sciscore}

Tables~\ref{tab:pref_acc_s} and~\ref{tab:pref_acc_h} break down the results from Table~\ref{tab:pref_acc} by individual task category. Abbreviations follow the definitions in Appendix~\ref{sec:supp_tasks}.~\model~achieves perfect accuracy (100\%) on multiple tasks across both splits, while its failures concentrate in subject-oriented tasks such as Buoyancy (BU) and Immiscibility (IM), where fine-grained subject-specific knowledge is required.
\begin{table*}[h]
  \centering
  \caption{\textbf{\model~achieves perfect accuracy on most tasks in~\dataset-S.} Per-category accuracy (\%). \colorbox{TabGreen}{Green} highlights the best VLM; \colorbox{TabBlue}{blue} highlights the best LMM.}
  \resizebox{0.98\linewidth}{!}{
  \begin{tabular}{l|cccccccccccccccc}
    \toprule
         Model & {ME} & {DI} & {EL} & {SO} & {IM} & {EV} & {AB} & {LI} & {FR} & {SC} & {RI} & {RU} & {LR} & {WR} & {BU} & {GR}\\
    \midrule
    CLIP-H~\cite{ilharco_gabriel_2021_5143773} & $25.00$ & $71.43$ & $47.62$ & $40.48$ & $54.17$ & $26.67$ & \cellcolor{TabGreen}$57.14$ & \cellcolor{TabGreen}$77.78$ & $73.33$ & $81.48$ & $34.62$ & $16.67$ & $62.22$ & $31.11$ & \cellcolor{TabGreen}$63.89$ & \cellcolor{TabGreen}$78.33$ \\
    BLIPScore~\cite{li2022blipbootstrappinglanguageimagepretraining} & \cellcolor{TabGreen}$56.94$ & $50.00$ & \cellcolor{TabGreen}$52.38$ & $44.05$ & $53.12$ & $20.00$ & $38.10$ & $33.33$ & \cellcolor{TabGreen}$76.67$ & $58.33$ & $38.46$ & \cellcolor{TabGreen}$42.86$ & \cellcolor{TabGreen}$76.67$ & \cellcolor{TabGreen}$38.89$ & $50.00$ & $47.50$ \\
    SigLIP ViT-SO-14~\cite{zhai2023sigmoidlosslanguageimage} & $44.44$ & \cellcolor{TabGreen}$83.33$ & $47.62$ & \cellcolor{TabGreen}$45.24$ & \cellcolor{TabGreen}$60.42$ & \cellcolor{TabGreen}$63.33$ & \cellcolor{TabGreen}$57.14$ & $58.33$ & $62.22$ & \cellcolor{TabGreen}$83.33$ & \cellcolor{TabGreen}$53.85$ & $23.81$ & $46.67$ & $33.33$ & $58.33$ & \cellcolor{TabGreen}$78.33$ \\
    \midrule
    LLaVA-OV-7B~\cite{li2024llavaonevisioneasyvisualtask} & $36.11$ & $75.00$ & $77.38$ & \cellcolor{TabBlue}$55.95$ & $55.21$ & \cellcolor{TabBlue} \textbf{100.00} & $38.10$ & \cellcolor{TabBlue}$95.83$ & $48.89$ & $78.70$ & $46.15$ & $59.52$ & $51.11$ & $72.22$ & $45.83$ & $92.50$\\
    Qwen2-VL-7B~\cite{wang2024qwen2vlenhancingvisionlanguagemodels} & $26.39$ & $70.24$ & $77.38$ & $42.86$ & $66.67$ & $68.33$ & $40.48$ & $84.72$ & $52.22$ & $95.37$ & $34.62$ & $73.81$ & $57.78$ & $67.78$ & $47.22$ & $83.33$\\
    InternVL2.5-8B~\cite{chen2024internvlscalingvisionfoundation} & \cellcolor{TabBlue}$41.67$ & $63.10$ & $72.62$ & $52.38$ & $56.25$ & $91.67$ & \cellcolor{TabBlue}$47.62$ & $90.28$ & $52.22$ & $84.26$ & \cellcolor{TabBlue}$75.00$ & $69.05$ & $84.44$ & \cellcolor{TabBlue}$90.00$ & $55.56$ & \cellcolor{TabBlue}$96.67$\\
    GPT-4o mini & $36.11$ & $77.38$ & $82.14$ & $35.71$ & $65.63$ & \cellcolor{TabBlue}\textbf{100.00} & $33.33$ & $76.39$ & \cellcolor{TabBlue}$58.89$ & $97.22$ & $53.85$ & $95.24$ & \cellcolor{TabBlue}$96.67$ & $83.33$ & \cellcolor{TabBlue}$56.94$ & $71.31$ \\
    GPT-4o mini$+$~CoT~\cite{wei2023chainofthoughtpromptingelicitsreasoning} & $36.11$& \cellcolor{TabBlue}$85.71$ & \cellcolor{TabBlue}$86.90$ & $45.24$ & \cellcolor{TabBlue}$68.75$ & \cellcolor{TabBlue}\textbf{100.00} & $33.33$ & $81.94$ & $56.67$ & \cellcolor{TabBlue}$98.15$ & $61.54$ & \cellcolor{TabBlue}$97.62$ & \cellcolor{TabBlue}$96.67$ & $88.89$ & $52.78$ & $80.33$\\
    \midrule
    Human Eval & $98.15$ & $65.87$ & $95.63$ & $86.11$ & \textbf{77.78} & \textbf{100.00} & $66.67$ & $82.08$ & $80.95$ & $90.74$ & $94.62$ & $92.86$ & $96.89$ & $99.56$ & \textbf{74.55} & $92.99$\\
    \model~(ours) & \textbf{100.00} & \textbf{97.62} & \textbf{100.00} & \textbf{90.48} & $68.75$ & \textbf{100.00} & \textbf{71.43} & \textbf{100.00} & \textbf{97.78} & \textbf{100.00} & \textbf{100.00} & \textbf{100.00} & \textbf{100.00} & \textbf{100.00} & $66.67$ & \textbf{98.33} \\
    \bottomrule
  \end{tabular}
  }
    
  \label{tab:pref_acc_s}
\end{table*}

\begin{table*}[h]
  \centering
  \caption{\textbf{\model~maintains strong per-category performance on~\dataset-C despite complex backgrounds.} Per-category accuracy (\%). \colorbox{TabGreen}{Green} highlights the best VLM; \colorbox{TabBlue}{blue} highlights the best LMM.}
\resizebox{0.98\linewidth}{!}{
  \begin{tabular}{l|cccccccccccccccc}
    \toprule
         Model & {ME} & {DI} & {EL} & {SO} & {IM} & {EV} & {AB} & {LI} & {FR} & {SC} & {RI} & {RU} & {LR} & {WR} & {BU} & {GR}\\
    \midrule
    CLIP-H~\cite{ilharco_gabriel_2021_5143773} & \cellcolor{TabGreen}$66.67$ & $78.57$ & $21.43$ & \cellcolor{TabGreen}$57.14$ & $50.00$ & $0.00$ & \cellcolor{TabGreen}$64.29$ & \cellcolor{TabGreen}$66.67$ & $46.67$ & $88.89$ & $75.00$ & $35.71$ & \cellcolor{TabGreen}$80.00$ & \cellcolor{TabGreen}$60.00$ & $58.33$ & $75.00$ \\
    BLIPScore~\cite{li2022blipbootstrappinglanguageimagepretraining} & $58.33$ & $50.00$ & $28.57$ & $42.86$ & \cellcolor{TabGreen}$62.50$ & \cellcolor{TabGreen}$50.00$ & $50.00$ & $29.17$ & \cellcolor{TabGreen}$60.00$ & $75.00$ & $54.17$ & \cellcolor{TabGreen}$57.14$ & $53.33$ & $46.67$ & $62.50$ & $40.00$ \\
    SigLIP ViT-SO-14~\cite{zhai2023sigmoidlosslanguageimage} & $58.33$ & \cellcolor{TabGreen}$85.71$ & \cellcolor{TabGreen}$42.86$ & $42.86$ & \cellcolor{TabGreen}$62.50$ & $30.00$ & \cellcolor{TabGreen}$64.29$ & $41.67$ & \cellcolor{TabGreen}$60.00$ & \cellcolor{TabGreen}$94.44$ & \cellcolor{TabGreen}$83.33$ & $28.57$ & $46.67$ & $53.33$ & \cellcolor{TabGreen}$66.67$ & \cellcolor{TabGreen}$95.00$ \\
    \midrule
    LLaVA-OV-7B~\cite{li2024llavaonevisioneasyvisualtask} & $45.83$ & $71.43$ & $64.29$ & \cellcolor{TabBlue}$71.43$ & $59.38$ & \cellcolor{TabBlue}\textbf{100.00} & \cellcolor{TabBlue}$57.14$ & \cellcolor{TabBlue}$79.17$ & $36.67$ & $77.78$ & $79.17$ & $53.57$ & $63.33$ & $86.67$ & \cellcolor{TabBlue}$75.00$ & \cellcolor{TabBlue}\textbf{100.00}\\
    Qwen2-VL-7B~\cite{wang2024qwen2vlenhancingvisionlanguagemodels} & $41.67$ & $64.29$ & $67.86$ & $57.14$ & $53.13$ & $55.00$ & \cellcolor{TabBlue}$57.14$ & $70.83$ & $36.67$ & \cellcolor{TabBlue}$94.44$ & $75.00$ & $60.71$ & $93.33$ & \cellcolor{TabBlue}$96.67$ & $66.67$ & \cellcolor{TabBlue}\textbf{100.00}\\
    InternVL2.5-8B~\cite{chen2024internvlscalingvisionfoundation} & $58.33$ & \cellcolor{TabBlue}\textbf{92.86} & $57.14$ & $53.57$ & \cellcolor{TabBlue}$71.88$ & $95.00$ & \cellcolor{TabBlue}$57.14$ & $70.83$ & $50.00$ & $75.00$ & \cellcolor{TabBlue}$87.50$ & $75.00$ & $90.00$ & $93.33$ & \cellcolor{TabBlue}$75.00$ & $97.50$\\
    GPT-4o mini & \cellcolor{TabBlue}$67.65$ & $67.86$ & $64.29$ & $50.00$ & $68.75$ & $90.00$ & $50.00$ & $75.00$ & \cellcolor{TabBlue}$53.33$ & $88.89$ & \cellcolor{TabBlue}$87.50$ & $89.29$ & \cellcolor{TabBlue}\textbf{100.00} & $83.33$ & $54.17$ & $97.50$\\
    GPT-4o mini$+$~CoT~\cite{wei2023chainofthoughtpromptingelicitsreasoning} & \cellcolor{TabBlue}$67.65$ & $85.71$ & \cellcolor{TabBlue}$85.71$ & $57.14$ & $68.75$ & $95.00$ & $32.14$ & \cellcolor{TabBlue}$79.17$ & $50.00$ & $88.89$ & \cellcolor{TabBlue}$87.50$ & \cellcolor{TabBlue}\textbf{92.86} & \cellcolor{TabBlue}\textbf{100.00} & $93.33$ & $41.67$ & \cellcolor{TabBlue}\textbf{100.00} \\
    \midrule
    Human Eval & $91.03$ & $66.75$ & \textbf{90.87} & $77.55$ & \textbf{86.61} & $95.71$ & \textbf{78.57} & $76.79$ & $77.14$ & $96.83$ & $83.78$ & \textbf{92.86} & $88.57$ & $84.76$ & \textbf{83.33} & $98.57$ \\
    \model~(ours) & \textbf{100.00} & $85.71$ & $85.71$ & \textbf{92.86} & $81.25$ & \textbf{100.00} & $71.43$ & \textbf{100.00} & \textbf{100.00} & \textbf{100.00} & \textbf{100.00} & \textbf{92.86} & \textbf{100.00} & \textbf{100.00} & $41.67$ & \textbf{100.00} \\
    \bottomrule
  \end{tabular}
  }
  \label{tab:pref_acc_h}
\end{table*}

\section{Analysis of VLM, LMM, and Human Performance}
\label{sec:supp_analysis_sciscore}

We provide further analysis of the results in Section~\ref{sec:rw_result}.

\begin{figure}[h]
    \centering
    \begin{minipage}[b]{0.33\linewidth}
        \includegraphics[width=\linewidth]{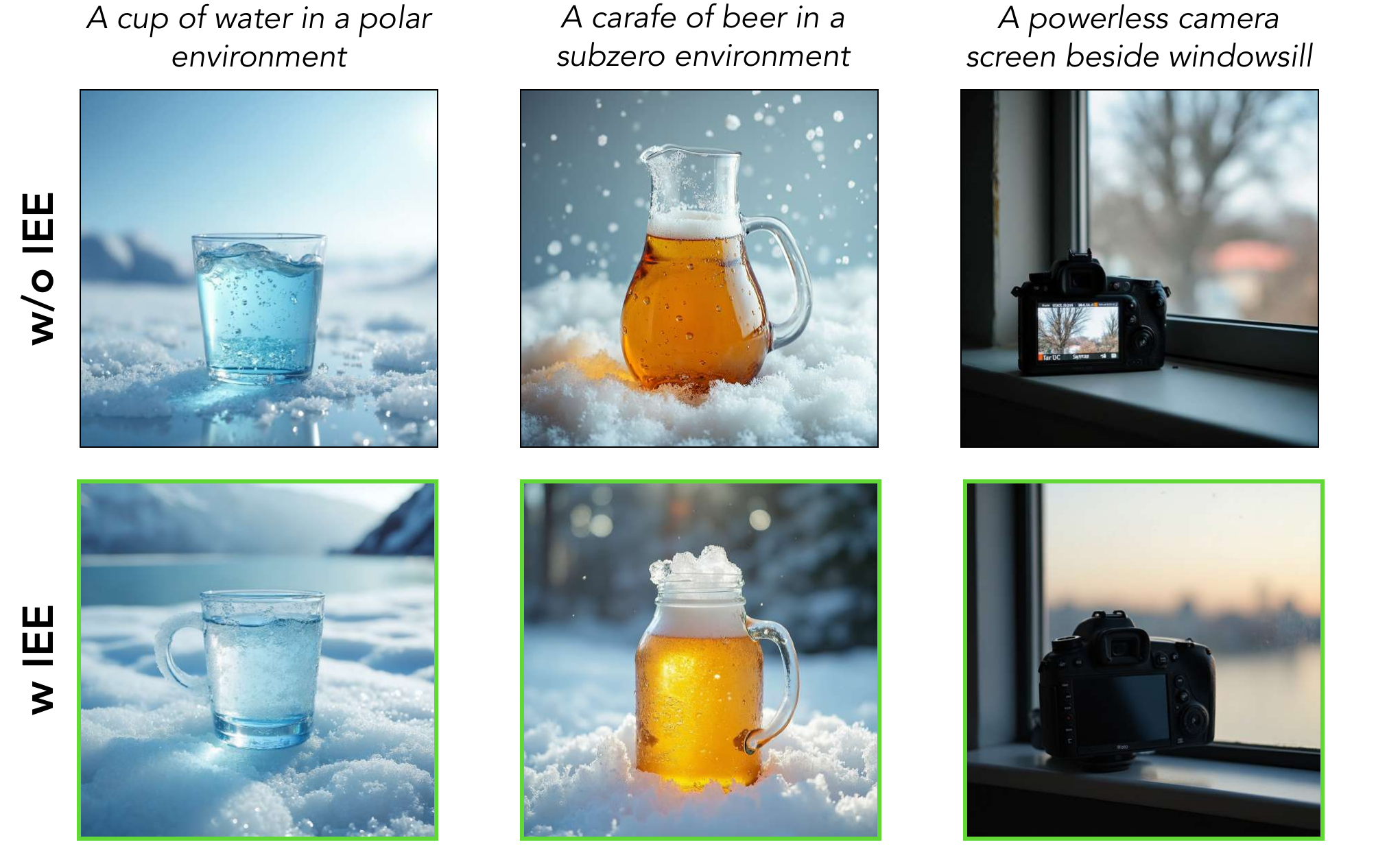}
        \caption{\textbf{Qualitative Results of IEE.} Images enclosed by green borders denote the correct selection in each pair.}
        \label{fig:IEE}
    \end{minipage}
    \hfill
    \begin{minipage}[b]{0.66\linewidth}
        \includegraphics[width=\linewidth]{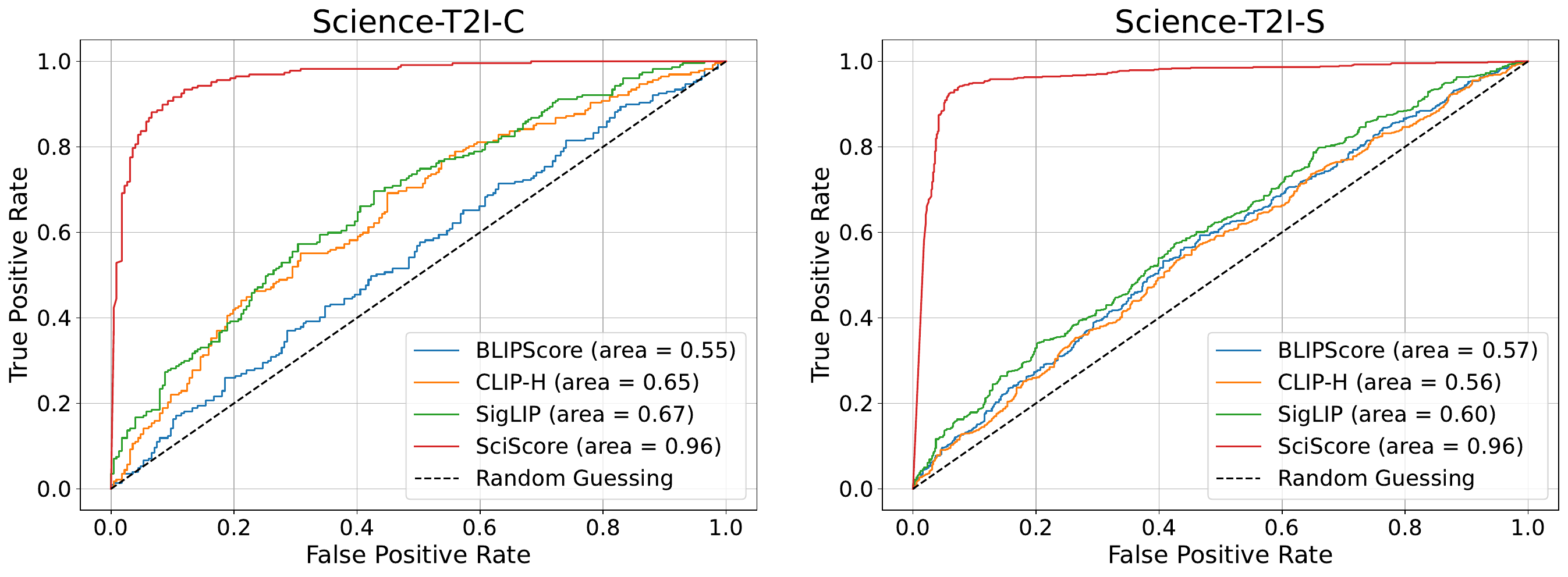}
        \caption{\textbf{\model~achieves near-perfect AUC while VLMs hover near 0.5.} ROC curves for CLIP-H, BLIPScore, SigLIP, and~\model.}
        \label{fig:roc}
    \end{minipage}
\end{figure}

\paragraph{VLM performance approaches random guessing.}
As shown by the ROC curves in Figure~\ref{fig:roc}, the AUC scores of CLIP-H~\cite{ilharco_gabriel_2021_5143773}, BLIPScore~\cite{li2022blipbootstrappinglanguageimagepretraining}, and SigLIP~\cite{zhai2023sigmoidlosslanguageimage} are only marginally above 0.5, confirming near-random performance. We attribute this to the nature of CLIP-style pretraining: the text encoder learns to match descriptive surface-level terms with visual content. When presented with an implicit prompt and two images that both contain the described subject, the model cannot distinguish between the scientifically correct and incorrect versions because the distinguishing features are not captured by the text-image alignment objective. In contrast,~\model~achieves a near-perfect AUC, demonstrating strong discriminative power.

\begin{figure*}[t]
    \centering
    \begin{subfigure}{\textwidth}
        \centering
        \begin{minipage}[b]{0.48\textwidth}
            \centering
            \includegraphics[width=\textwidth]{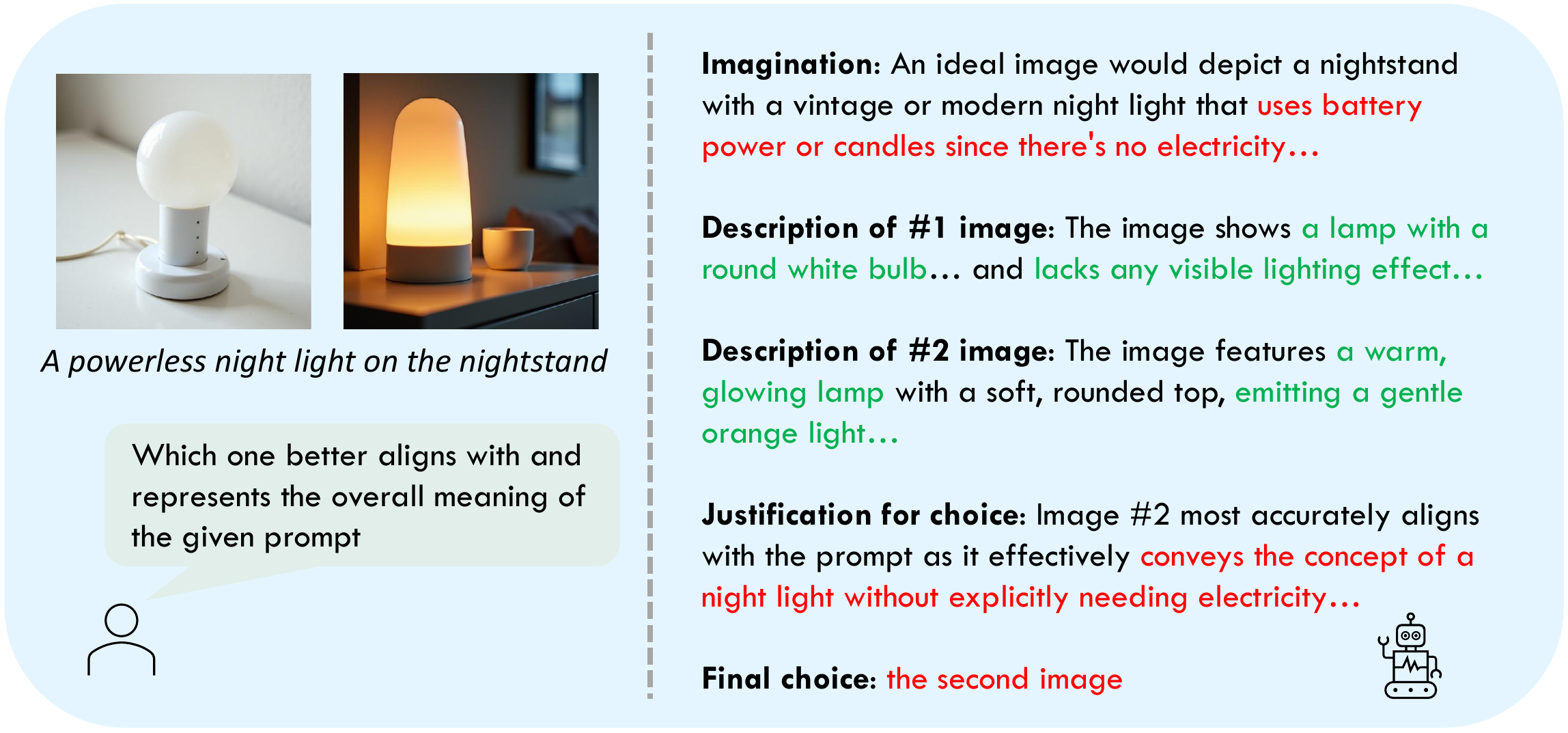}
        \end{minipage}
        \hfill
        \begin{minipage}[b]{0.48\textwidth}
            \centering
            \includegraphics[width=\textwidth]{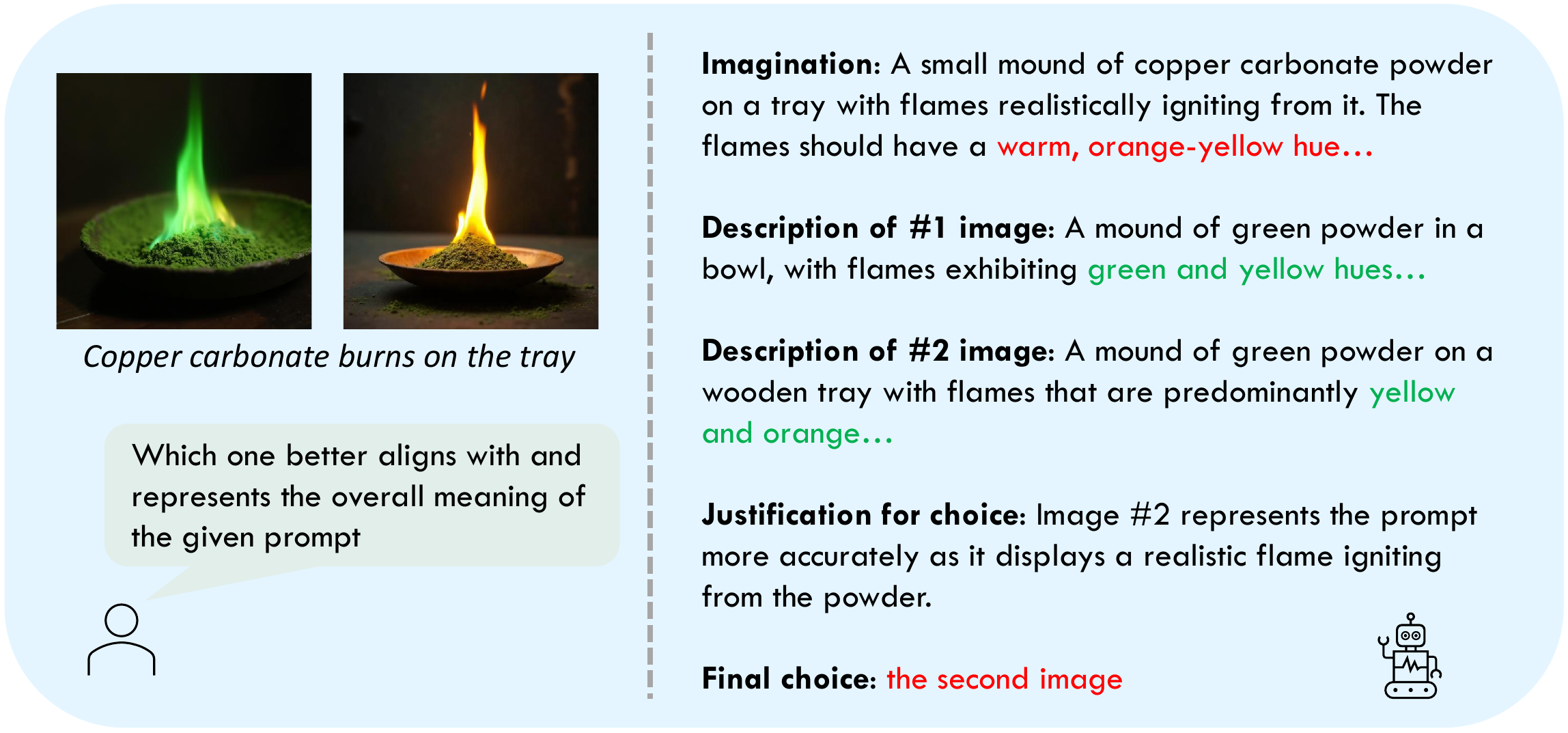}
        \end{minipage}
        
        \vspace{1em}
        \begin{minipage}[b]{0.48\textwidth}
            \centering
            \includegraphics[width=\textwidth]{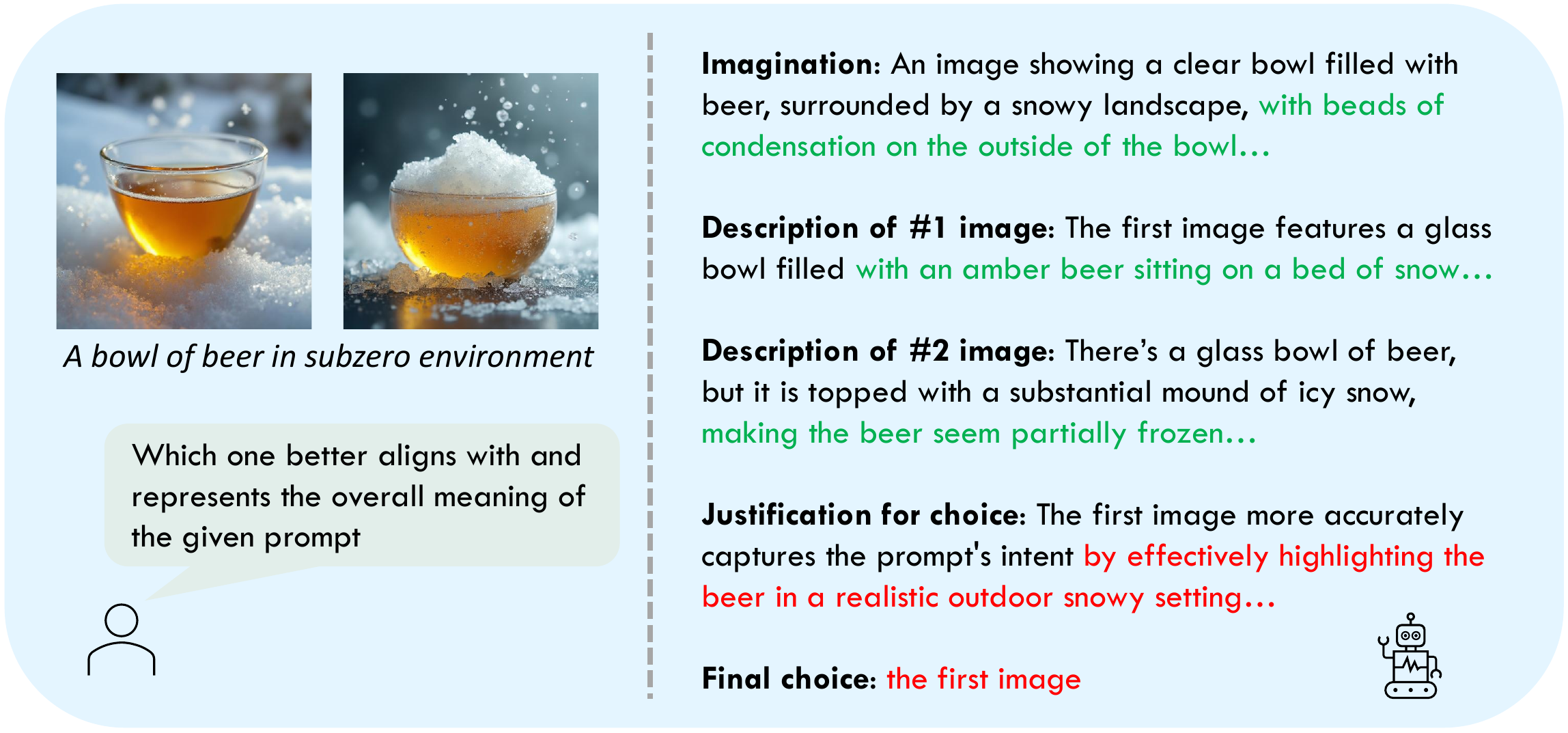}
        \end{minipage}
        \hfill
        \begin{minipage}[b]{0.48\textwidth}
            \centering
            \includegraphics[width=\textwidth]{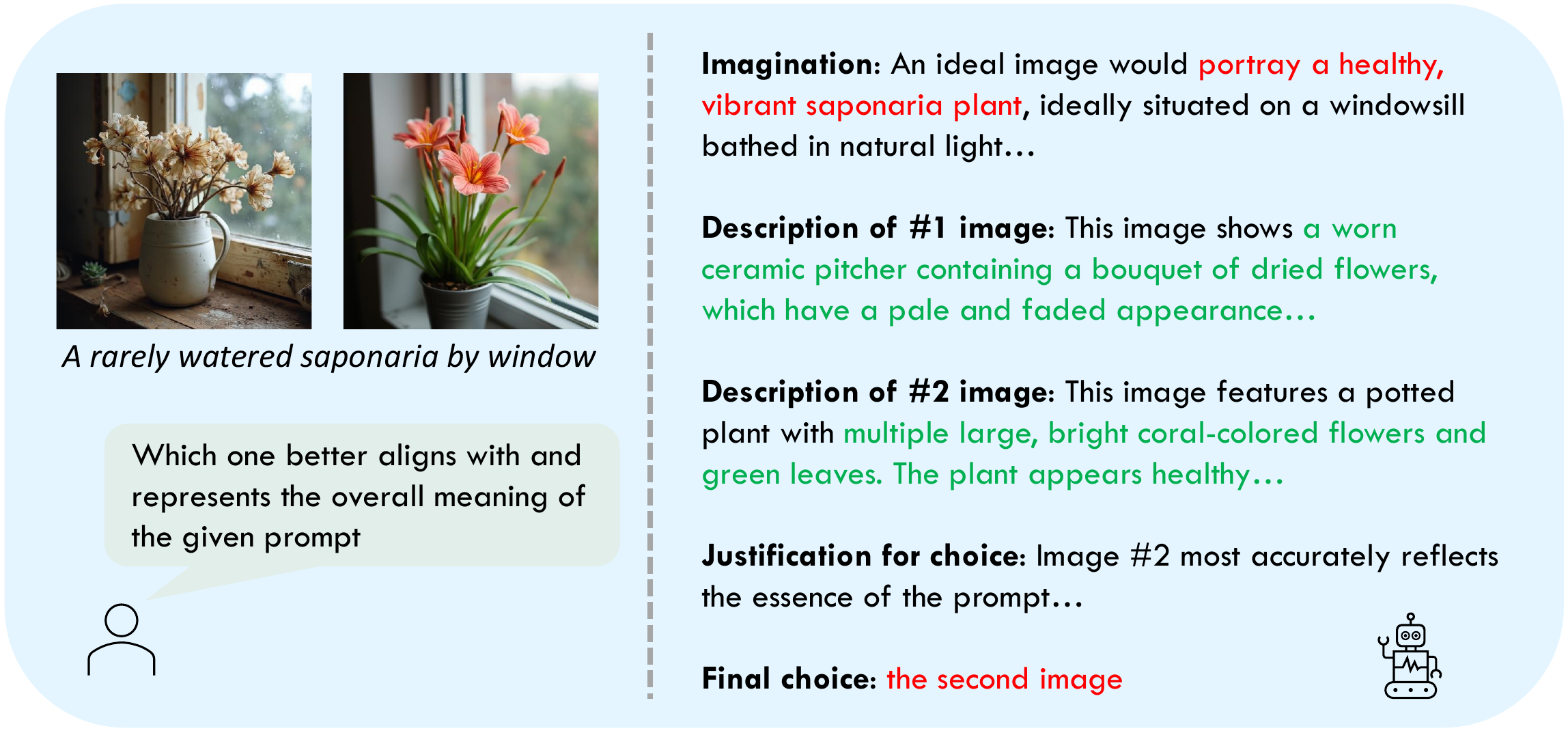}
        \end{minipage}
        \caption{\textbf{Reasoning Failure.} \texttt{GPT-4o-mini}~\cite{GPT} inaccurately infers the target image by misinterpreting the input prompt and neglecting the underlying scientific principles embedded within it. Instead of employing a systematic reasoning process, it relies predominantly on intuitive imagination.}
    \end{subfigure}
    
    \vspace{2em}
    
    \begin{subfigure}{\textwidth}
        \centering
        \begin{minipage}[b]{0.48\textwidth}
            \centering
            \includegraphics[width=\textwidth]{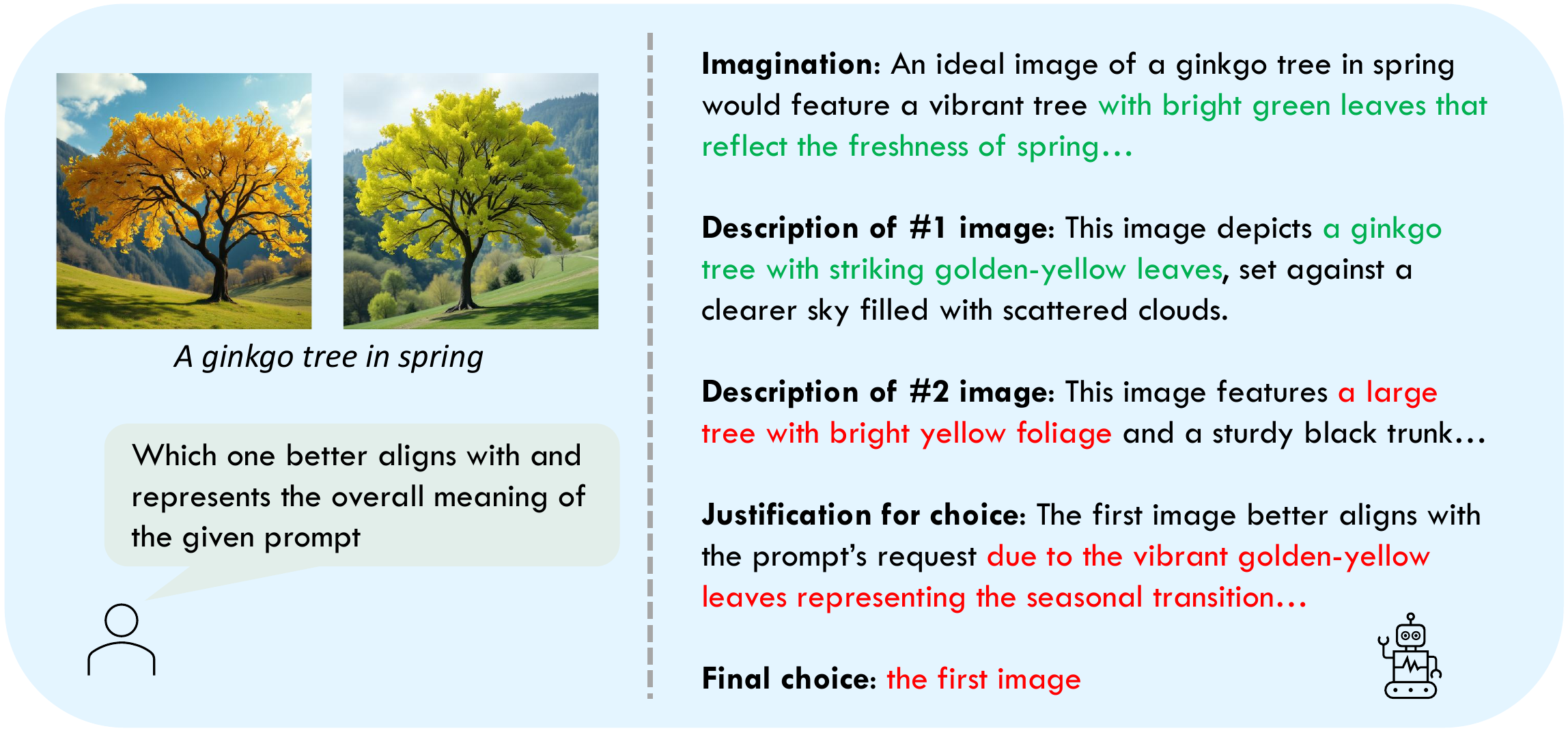}
        \end{minipage}
        \hfill
        \begin{minipage}[b]{0.48\textwidth}
            \centering
            \includegraphics[width=\textwidth]{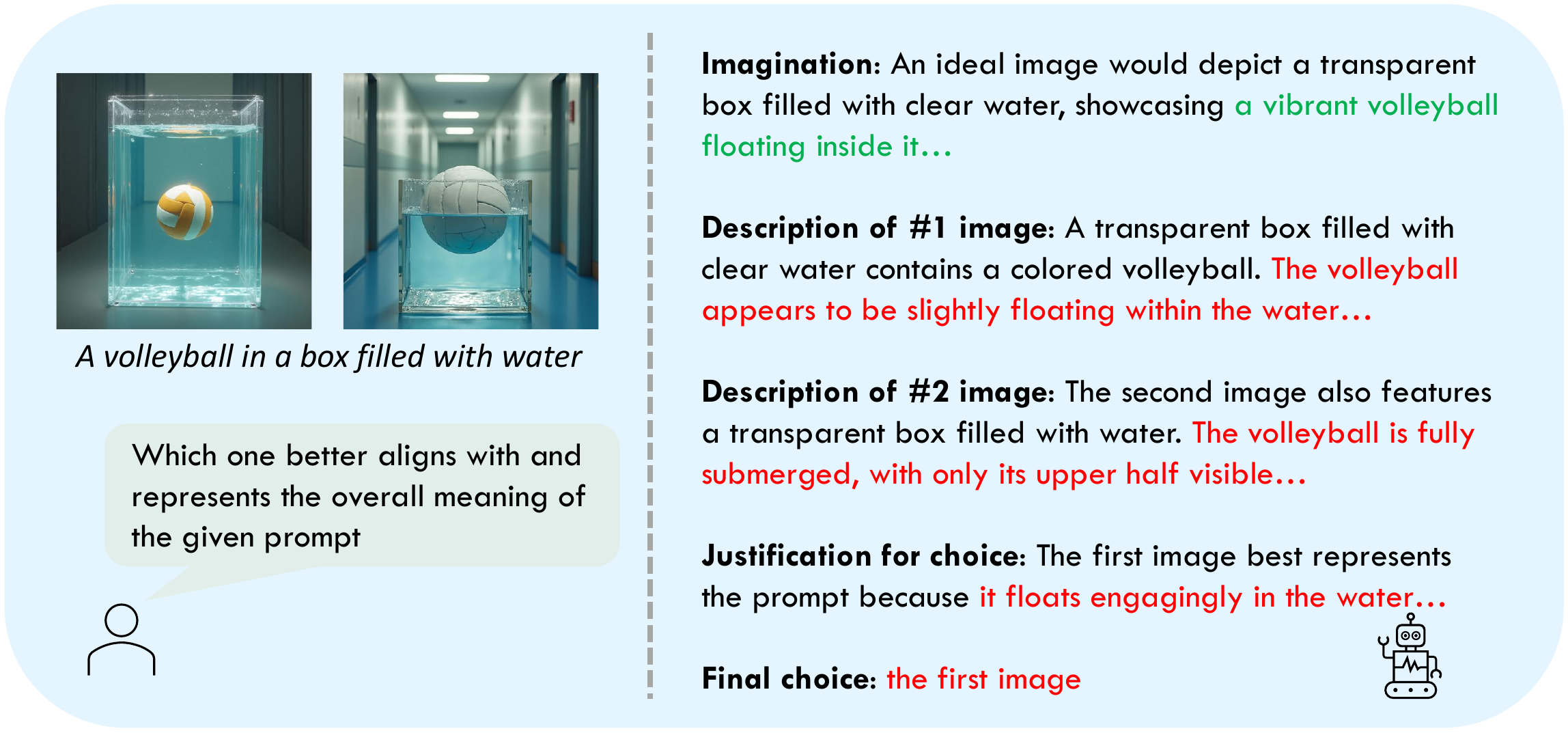}
        \end{minipage}
        
        \vspace{1em}
        \begin{minipage}[b]{0.48\textwidth}
            \centering
            \includegraphics[width=\textwidth]{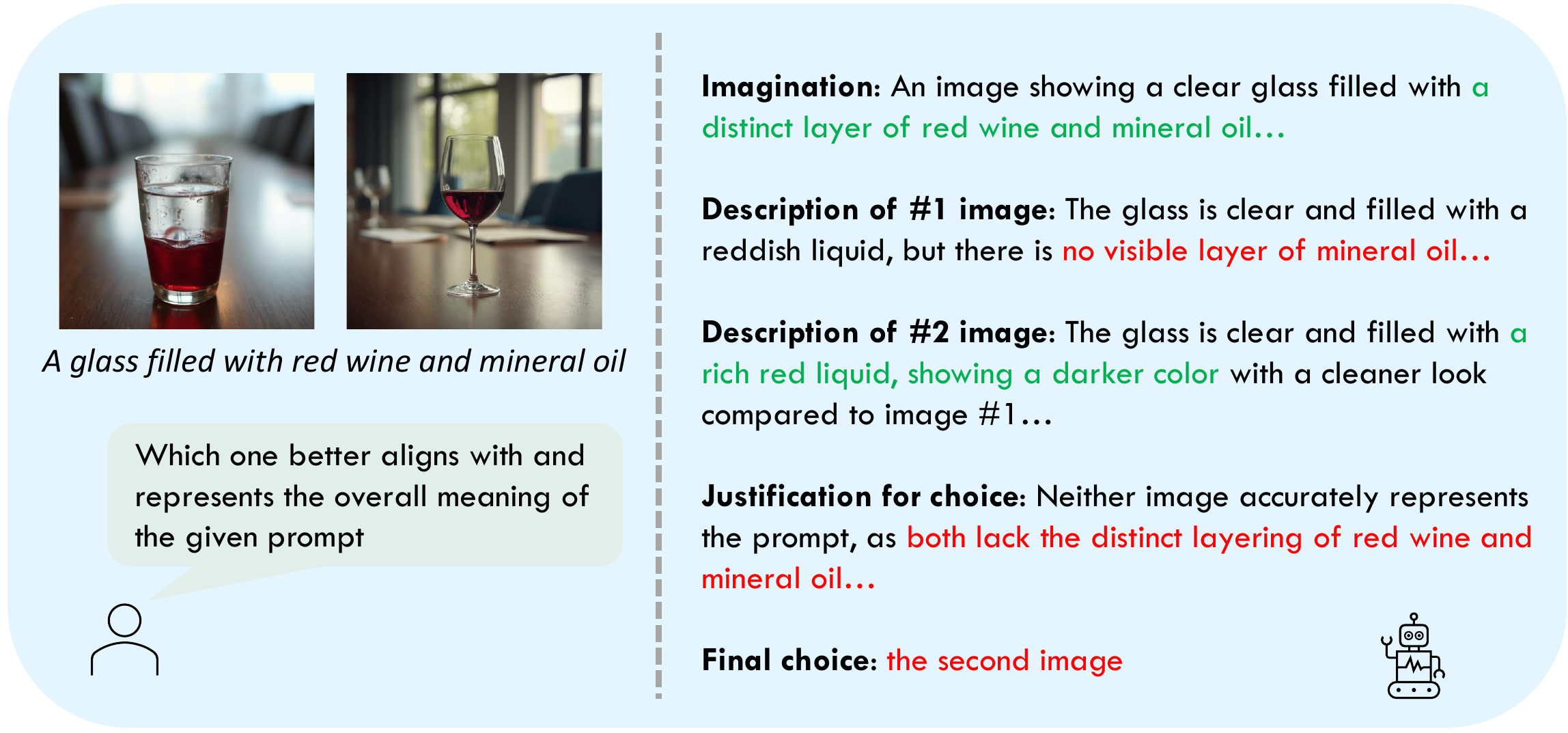}
        \end{minipage}
        \hfill
        \begin{minipage}[b]{0.48\textwidth}
            \centering
            \includegraphics[width=\textwidth]{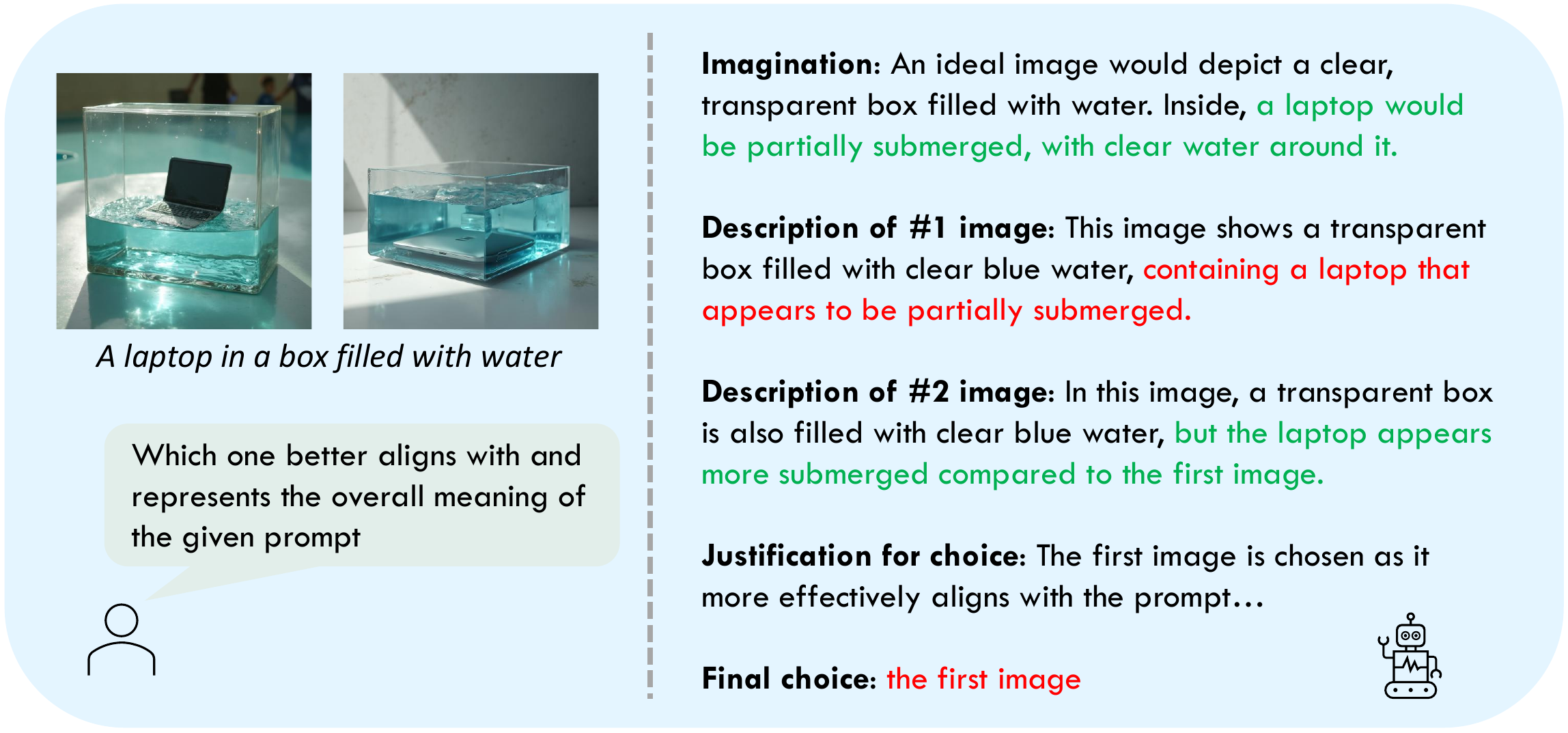}
        \end{minipage}
        \caption{\textbf{Visual Limitation.} \texttt{GPT-4o-mini}~\cite{GPT} inaccurately describes the image, thereby impeding the reasoning process. Specifically, for tasks involving spatial relationships, it fails to make correct judgments, resulting in erroneous interpretations of positional dynamics within the visual content.}
    \end{subfigure}
    \caption{\textbf{Qualitative Failure Cases of GPT}. In both cases, the CoT~\cite{wei2023chainofthoughtpromptingelicitsreasoning} reasoning approach from Figure~\ref{fig:gpt_eval} is applied, but errors in either interpretation or visual comprehension impact the final decision. Green text indicates correct inference, while red text marks errors.}
\label{fig:gpt_error}
\vspace{-2pt}
\end{figure*}

\paragraph{LMMs struggle with visual scientific reasoning.}
Despite access to extensive knowledge bases, all LMMs including \texttt{GPT-4o-mini}~\cite{GPT} underperform on our benchmark, even with CoT prompting~\cite{wei2023chainofthoughtpromptingelicitsreasoning}. We identify two primary failure modes, illustrated in Figure~\ref{fig:gpt_error}. First, the models often misinterpret fine-grained visual features critical to scientific tasks, such as subtle color differences or spatial relationships between objects. Second, even when visual perception is adequate, the models produce internally contradictory reasoning chains, undermining the reliability of their final judgments. These observations suggest that the bottleneck lies in visual perception and reasoning coherence rather than knowledge availability.

\paragraph{Why~\model~surpasses human evaluators.}
Human evaluators possess broad but finite scientific knowledge. Their expertise is inevitably bounded by their specific training backgrounds, which can lead to errors on tasks outside their immediate domain.~\model, by contrast, acquires task-specific knowledge from the~\dataset~training set and applies it consistently across all domains, explaining its roughly 5\% advantage over human evaluators.

\section{Qualitative Analysis of IEE} 
\label{sec:supp_abla_sciscore}

Figure~\ref{fig:IEE} illustrates the effect of IEE on~\model's ability to capture fine-grained visual details. The first two examples involve distinguishing between frozen and liquid states, which requires detecting subtle differences in transparency. The third example requires identifying whether a small screen region displays meaningful content. Without IEE ($\lambda = 0$), the model fails on these cases because the image encoder lacks sensitivity to such fine-grained differences. With IEE ($\lambda = 0.25$), the enhanced image encoder correctly identifies the distinguishing features, confirming that IEE strengthens the model's visual discrimination at the level required for scientific assessment.

\section{Details of Two-Stage Training}
\label{sec:supp_two_stage_train_detail}
We provide the full mathematical derivations underlying our two-stage alignment framework.
\paragraph{Supervised Fine-tuning (SFT).}
Flow matching models \cite{lipman2023flowmatchinggenerativemodeling} are continuous-time generative models that define a time-dependent velocity field $v(x_t, t)$ to transport samples from a noise distribution $p_1$ to data distribution $p_0$ over a time interval $t \in [0, 1]$. The transformation is governed by the ordinary differential equation (ODE):
\begin{equation}
    \frac{dx_t}{dt} = v(x_t, t),
\end{equation}
with the initial condition $ x_1 \sim p_1 $. The forward process is constructed as:
\begin{equation}
  x_t = \alpha_t x_0 + \beta_t \epsilon, \quad \epsilon \sim \mathcal{N}(0, I),
\end{equation}
where $\alpha_0 = 1$, $\beta_0 = 0$, $\alpha_1 = 0$, and $\beta_1 = 1$, ensuring the consistency of the marginal distributions with the initial and terminal conditions. The velocity field $v(x_t, t)$ is represented as the sum of two conditional expectations:
\begin{equation}
    v(x, t) = \dot{\alpha}_t \mathbb{E}[x_* | x_t = x] + \dot{\beta}_t \mathbb{E}[\epsilon | x_t = x],
\end{equation}
which can be approximated by the model $v_\theta(x, t)$ by minimizing the following training objective:
\begin{equation}
    \mathcal{L}_{SFT}(\theta) := \mathbb{E}_{{x}_*, \boldsymbol{\epsilon}, t} \left[ \| {v}_\theta({x}_t, t) - \dot{\alpha}_t {x}_* - \dot{\beta}_t \epsilon \|^2 \right]
\end{equation}

\paragraph{Direct Preference Optimization (DPO).} RLHF aims to optimize a conditional distribution 
$p_\theta(x_0|c)$ such that the expected reward $r(c, x_0)$ is maximized, while simultaneously regularizing 
the KL-divergence from a reference distribution $p_{\text{ref}}$. This objective is formulated as:
\begin{equation}
\max_{p_\theta} \mathbb{E}_{c, x_0 \sim p_\theta(x_0|c)} 
\left[ r(c, x_0) \right] 
- \beta \mathcal{D}_{\text{KL}} \left[ p_\theta(x_0|c) \| p_{\text{ref}}(x_0|c) \right]
\end{equation}
where the hyper-parameter $\beta$ controls regularization. According to~\cite{rafailov2024directpreferenceoptimizationlanguage}, the unique global optimal solution $p_{\theta}^*$ to this optimization problem is given by:
\begin{equation}
    p_{\theta}^*(x_0|c) = p_{\text{ref}}(x_0|c) \exp\left(\frac{r(c, x_0)}{\beta}\right) / Z(c)
\end{equation}
where $Z(c) = \sum_{x_0} p_{\text{ref}}(x_0|c) \exp\left(\frac{r(c, x_0)}{\beta}\right)$ is the partition function. Then the reward function can be expressed as:
\begin{equation}
    r(c, x_0) = \beta \log \frac{p_{\theta}^*(x_0|c)}{p_{\text{ref}}(x_0|c)} + \beta \log Z(c)
\end{equation}
To model human preferences, the Bradley-Terry (BT) model is employed, which represents the probability of one outcome being preferred over another as: 
\begin{equation}
    p_{BT}(x_0^w \succ x_0^l | c) = \sigma(r(c, x_0^w) - r(c, x_0^l))
\end{equation}
where $\sigma$ is the sigmoid function, $x_0^w$ is the preferred outcome, and $x_0^l$ is the less preferred one. \(r(c, x_0)\) can be parameterized by a neural network \(\phi\) and estimated via maximum likelihood training for binary classification:
\begin{equation}
    L_{BT}(\phi) = \mathbb{E}_{c, x_0^w, x_0^l} \left[ \log \sigma \left( r_\phi(c, x_0^l) - r_\phi(c, x_0^w) \right) \right]
\end{equation}
By leveraging the relationship between the reward function and the optimal policy $p_\theta^*$, the DPO objective is derived as:
\begin{equation}
    \mathcal{L}_{\text{DPO}}(\theta) = -\mathbb{E}_{c, x_0^w, x_0^l}\Bigg[ \log \sigma \bigg(\beta \log \frac{p_{\theta}(x_0^w|c)}{p_{\text{ref}}(x_0^w|c)} - \beta \log \frac{p_{\theta}(x_0^l|c)}{p_{\text{ref}}(x_0^l|c)} \bigg) \Bigg]
\end{equation}

\paragraph{Choice of $\sigma_t$.}

We determine the value of $\sigma_t$ by adhering to the methodology presented in~\cite{karras2022elucidatingdesignspacediffusionbased}. Initially, we define the hyperparameters $S_{\text{churn}}$, $S_{\min}$, $S_{\max}$, and $S_{\text{noise}}$. Subsequently, we define $\gamma_t$ as follows:
\begin{equation} 
\gamma_t = 
\begin{cases} 
\min\left(S_{\text{churn}} \cdot \Delta t, \sqrt{2} - 1\right) & \text{if } t \in [S_{\min}, S_{\max}]\\ 
0 & \text{otherwise}, 
\end{cases}
\end{equation}
where $\Delta t$ represents the timestep difference between consecutive sampling steps. Following this, we define $\sigma_t$ by
\begin{equation} 
\sigma_t = S_{\text{noise}} \cdot \sqrt{\gamma_t^2 + 2\gamma_t} \cdot (1 - t). \end{equation}

\paragraph{Subject extraction.} Before training begins, we use an LLM to extract the subject from each prompt. During OFT, the extracted subjects are passed to GroundingDINO~\cite{liu2024groundingdinomarryingdino} to produce bounding-box masks for each generated image.

\paragraph{Gradient masking in latent space.} GroundingDINO produces masks at RGB resolution, but gradients are computed in the model's latent space. Because the pretrained VAE~\cite{kingma2022autoencodingvariationalbayes} used by LDM~\cite{rombach2022highresolutionimagesynthesislatent} preserves spatial locality, we map the bounding box directly. Let the latent have dimensions $(H_l, W_l, C_l)$ and the decoded image $(H, W, C)$. A bounding box $(x_1, y_1, x_2, y_2)$ in pixel space maps to:
\begin{equation}
    \left( \frac{x_1}{H} \cdot H_l,\ \frac{y_1}{W} \cdot W_l,\ \frac{x_2}{H} \cdot H_l,\ \frac{y_2}{W} \cdot W_l \right)
\end{equation}
This latent-space mask is applied to the gradients during backpropagation.

\paragraph{Mask padding.} For tasks where positional relationships matter (e.g., an object's height relative to the ground in the \textit{gravity} task), the tight bounding box around the subject is insufficient. We therefore pad the mask by 10\% in both height and width to capture the surrounding spatial context.
\section{Two-Stage Training Settings}
\label{sec:supp_ft_train_setting}
Table~\ref{tab:t2i_hyper} lists the hyper-parameter configurations for both SFT and OFT stages.

\begin{table}[t]
    \centering
    \caption{\textbf{Hyper-parameters for the two-stage alignment framework.}}
    \begin{tabular}{l|c|c}
        \toprule
        Hyper-parameters & SFT & OFT\\
        \midrule
        batch size & $32$ & $8$ \\
        learning rate & $2 \times 10^{-5}$ & $3 \times 10^{-4}$ \\
        training steps & $900$ & $140$ \\
        optimizer & AdamW~\cite{loshchilov2019decoupledweightdecayregularization} & AdamW~\cite{loshchilov2019decoupledweightdecayregularization} \\
        gradient accumulation & $8$ & $2$ \\
        LoRA rank & $16$ & $16$ \\
        $S_{\text{churn}}$ & / & $0.1$ \\
        $S_{\min}, S_{\max}$ & / & $0, \infty$ \\
        $S_{\text{noise}}$ & / & $1.0$ \\
        $\beta$ & / & $10$ \\
        \bottomrule 
    \end{tabular}

    \label{tab:t2i_hyper}
\end{table}

\section{Additional Results of Online Fine-tuning}
\label{sec:supp_res_oft}
To verify that our OFT pipeline generalizes beyond~\model, we conduct additional experiments using two alternative reward models: the LAION aesthetic predictor~\cite{Laion-aesthetic} and ImageReward~\cite{xu2023imagereward}. In these experiments, we do not apply the masking strategy from the main text, as the tasks (animal generation) do not require subject-level localization.

\paragraph{Training.}
We fine-tune FLUX.1[schnell]~\cite{FLUX} with 4 inference steps. All other configurations follow Table~\ref{tab:t2i_hyper} except: 128 images per step, learning rate $6 \times 10^{-5}$, gradient accumulation of 8. We train for 164 steps with LAION and 550 steps with ImageReward. Following DDPO~\cite{black2024trainingdiffusionmodelsreinforcement}, the training set comprises 45 animal categories.

\begin{figure*}[t]
    \vspace{-3pt}
    \centering
    \includegraphics[width=\linewidth]{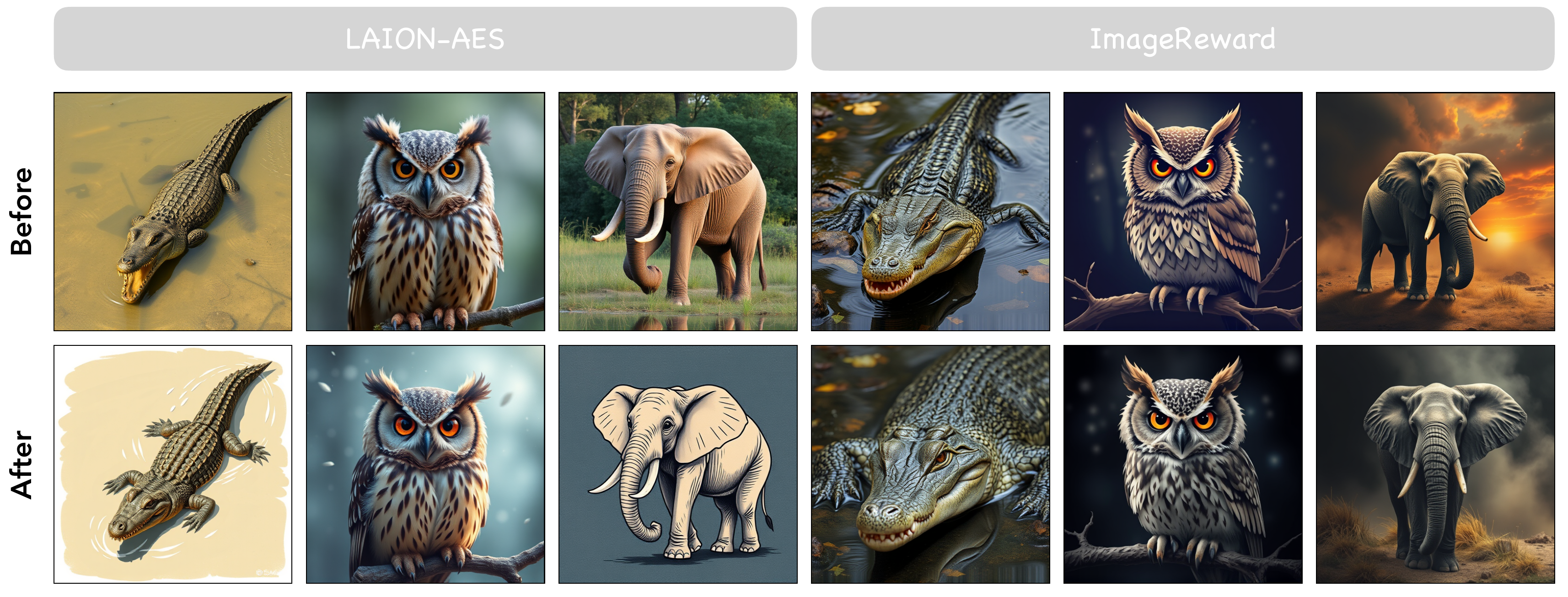}
    \caption{\textbf{OFT improves visual quality with alternative reward models.} Upper: base FLUX.1[schnell]~\cite{FLUX}. Lower: after OFT. Each pair uses an identical random seed. Prompts: ``crocodile'', ``owl'', ``elephant.''}
    \label{fig:case_study_aes}
\end{figure*}

\paragraph{Evaluation.}
The test set consists of 10 held-out animal categories. For each prompt, we generate 100 images and report the average reward. Results are shown in Table~\ref{tab:alg_res}; qualitative examples are provided in Figure~\ref{fig:case_study_aes}.

\begin{table}[h!]
    \centering
    \caption{\textbf{OFT consistently improves rewards with alternative reward models.} Average reward on 10 held-out animal categories.}
    \begin{tabular}{l|c|c}
        \toprule
        Method & LAION~\cite{Laion-aesthetic} & ImageReward~\cite{xu2023imagereward}\\
        \midrule
        FLUX.1[schnell]~\cite{FLUX} & $5.855$ & $0.949$ \\
        $+$OFT & \textbf{6.074} & \textbf{1.023} \\
        \bottomrule 
    \end{tabular}
    \label{tab:alg_res}
    \vspace{-6pt}
\end{table}

\section{Limitations and Future Directions}
\label{sec:supp_limitation}

\paragraph{Limited cross-task generalizability.}
Scientific knowledge does not transfer easily across domains: expertise in dynamics does not inherently facilitate reasoning about thermodynamics, even for humans. As a result, our alignment framework improves performance primarily on the 16 predefined tasks, and extending it to new scientific domains would require additional domain-specific training data.

\paragraph{Subject-oriented tasks remain difficult.}
As discussed in Section~\ref{sec:rw_result}, subject-oriented tasks pose the greatest challenge because they require knowledge of each subject's specific properties. When the model encounters a novel or unseen subject, it lacks the prior exposure needed to generate the correct visual features. Incorporating world knowledge from LLMs or scaling up the training data to cover a broader range of subjects are promising directions for future work.

\clearpage

\begin{figure}[t]
\centering
\begin{tcolorbox}[title=Grading Criteria Examples]

\textbf{Example 1}\\[3pt]
\textbf{Prompt:} ``A transparent water-filled box holds a basketball, depicted realistically."\\[3pt]
\textbf{Scene Grading:}
\begin{itemize}[leftmargin=12pt, itemsep=1pt, topsep=2pt]
    \item \textit{0 point:} The picture does not feature a basketball inside a transparent box filled with water in any capacity.
    \item \textit{1 point:} The picture shows a basketball, but it is not inside a transparent box. Alternatively, the basketball are in a transparent box, but there is no water present.
    \item \textit{2 points:} The picture accurately depicts a basketball inside a transparent box filled with water.
\end{itemize}
\textbf{Reality Grading:}
\begin{itemize}[leftmargin=12pt, itemsep=1pt, topsep=2pt]
    \item \textit{0 point:} The basketball is completely sinking to the bottom of the water.
    \item \textit{1--2 points:} The basketball is completely submerging in the water, but doesn't reach the bottom. Less mistakes will earn a higher score.
    \item \textit{3 points:} The picture shows basketball floating on the surface of the water.
\end{itemize}

\tcblower

\textbf{Example 2}\\[3pt]
\textbf{Prompt:} ``A clear glass filled with water and oil, simple and realistic."\\[3pt]
\textbf{Scene Grading:}
\begin{itemize}[leftmargin=12pt, itemsep=1pt, topsep=2pt]
    \item \textit{0 point:} There is no glass or no liquid in the glass, or the scene is irrelevant (e.g., the focus is not on the glass or liquid at all).
    \item \textit{1 point:} The glass contains liquid, but the focus on the liquid or the glass is unclear, or there are distracting elements in the scene.
    \item \textit{2 points:} The glass is clearly depicted with some liquid in it, with no distractions, offering a simple, clear, and realistic depiction.
\end{itemize}
\textbf{Reality Grading:}
\begin{itemize}[leftmargin=12pt, itemsep=1pt, topsep=2pt]
    \item \textit{0 points:} Liquids are mixed or incorrectly positioned (e.g., water and oil blended or misplaced).
    \item \textit{1 point:} Water and oil are present but with partial inaccuracies in separation or positioning (e.g., water floating on oil, blurred boundaries).
    \item \textit{2 points:} Liquids are correctly positioned with visible separation (oil atop water), but minor deviations from realism exist (e.g., slight issues with clarity or texture).
    \item \textit{3 points:} Fully realistic depiction with correct positioning (oil floating on water) and clear separation.
\end{itemize}
\end{tcolorbox}
\vspace{-6pt}
\caption{\textbf{Representative grading criteria for two tasks.} Each tuple specifies the Scene Score rubric (whether the main subject is present) and the Reality Score rubric (whether the scientific phenomenon is correctly depicted).}
\label{fig:sample_exp}
\vspace{-6pt}
\end{figure}

\begin{figure}[t]
\centering
\begin{tcolorbox}[title=Evaluation Instruction]
\small
You are an experienced scientist. Begin by evaluating the provided image using the specified
scene composition criteria. If the image does not fully satisfy these criteria, assign a reality score of
0. However, if the scene meets all the criteria, proceed to assess its realism based on the given reality
scoring guidelines, disregarding stylistic aspects and minor background details. Please first describe the
image in detail and then adhere strictly to these criteria to ensure an accurate scoring of the image.\\[6pt]
\textbf{Input:} \{``Prompt": [Your Input Prompt], ``Scene Grading": [Your Input Scene Grading], ``Reality Grading": [Your Input Reality Grading], ``Image": [Your Input Image]\}.\\[6pt]
\textbf{Output format:} \{``description":, ``scene score": , ``reality score": \}
\end{tcolorbox}
\vspace{-6pt}
\caption{\textbf{Instruction template for image evaluation.} The evaluator receives an image, its implicit prompt, and the corresponding grading criteria, then produces Scene Score (SS) and Reality Score (RS).}
\label{fig:user_inst}
\vspace{-6pt}
\end{figure}

\begin{figure*}[t]
    \centering
    \subfloat{
        \includegraphics[width=\linewidth]{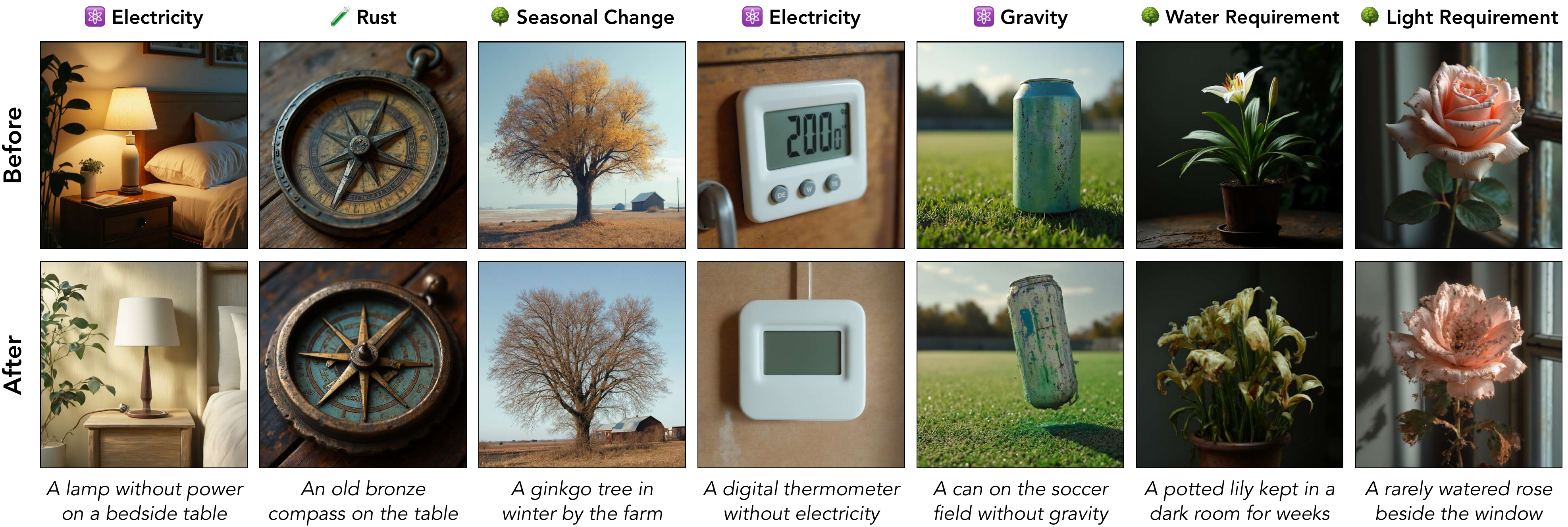}
    }
    \hfill
    \vspace{6pt}
    \subfloat{
        \includegraphics[width=\linewidth]{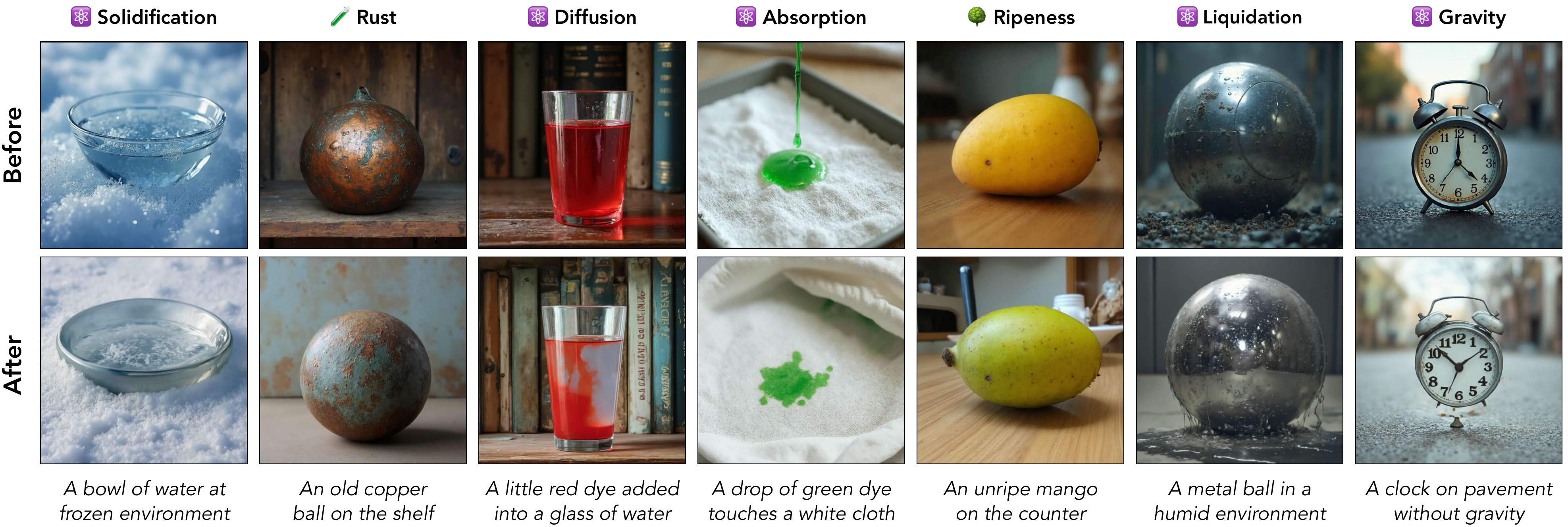}
    }
    \hfill
    \vspace{6pt}
    \subfloat{
        \includegraphics[width=\linewidth]{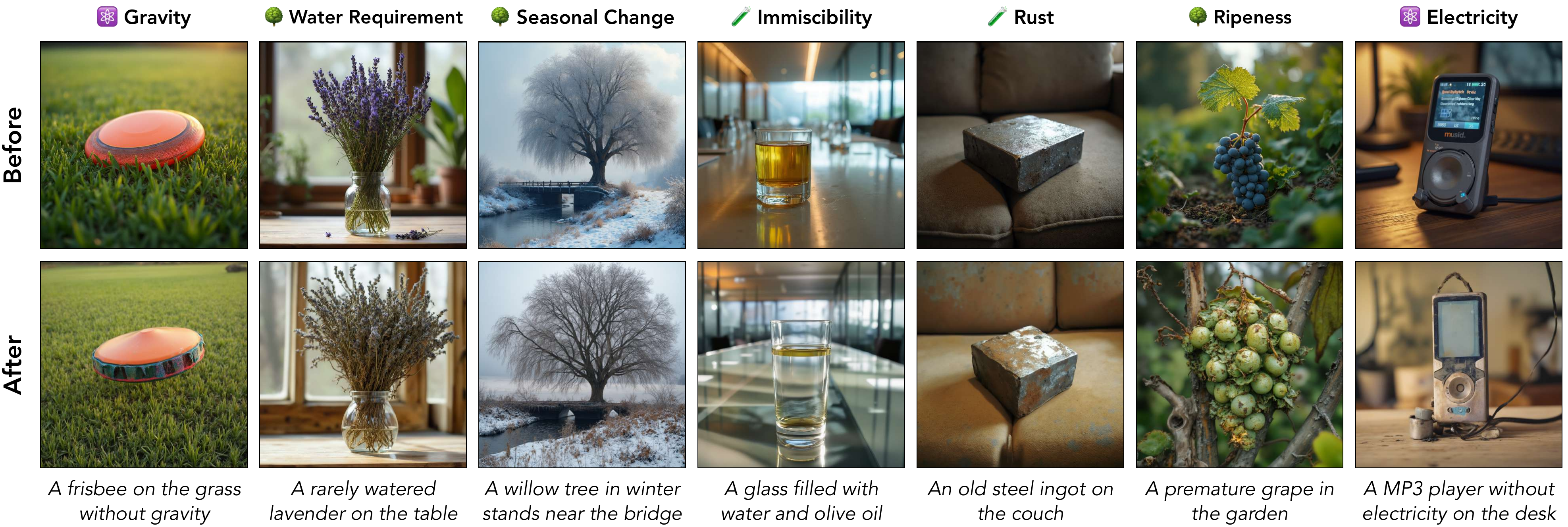}
    }
    \caption{\textbf{Additional Generated Samples.} Each pair of images is produced using the same random seed to ensure consistency.}
    \label{fig:generated_case}
\end{figure*}

\end{document}